\documentclass[sn-basic,authoryear]{sn-jnl}

\usepackage{threeparttable}
\usepackage{comment}
\usepackage{graphicx}%
\usepackage{multirow}%
\usepackage{amsmath,amssymb,amsfonts}%
\usepackage{amsthm}%
\usepackage{mathrsfs}%
\usepackage[title]{appendix}%
\usepackage{xcolor}%
\usepackage{textcomp}%
\usepackage{manyfoot}%
\usepackage{booktabs}%
\usepackage{algorithm}%
\usepackage{algorithmicx}%
\usepackage{algpseudocode}%
\usepackage{listings}%
\usepackage[normalem]{ulem}

\usepackage{float}
\usepackage{subfigure}
\usepackage{kotex}
\usepackage{tikz}
\usetikzlibrary{positioning}
\usepackage[section]{placeins}
\usepackage{hyperref}

\begin{document}

\title[Enhancing Social Media Post Popularity Prediction with Visual Content]{Enhancing Social Media Post Popularity Prediction with Visual Content}

\author[1]{Dahyun Jeong} \email{ekgus638@uos.ac.kr}

\author*[2]{Hyelim Son } \email{hlson@uos.ac.kr}

\author[1]{Yunjin Choi} \email{ycstat@uos.ac.kr}

\author[3]{Keunwoo Kim} \email{keunwoo.kim@uos.ac.kr}

\affil[1]{Department of Statistics, University of Seoul, Seoulsiripdae-ro 163}

\affil*[2]{School of Economics, University of Seoul, Seoulsiripdae-ro 163}

\affil[3]{College of Business Administration, University of Seoul, Seoulsiripdae-ro 163}
\abstract{
Our study presents a framework for predicting image-based social media content popularity that focuses on addressing complex image information and a hierarchical data structure.
We utilize the Google Cloud Vision API to effectively extract key image and color information from users' postings, achieving 6.8\% higher accuracy compared to using non-image covariates alone. For prediction, we explore a wide range of prediction models, including Linear Mixed Model, Support Vector Regression, Multi-layer Perceptron, Random Forest, and XGBoost, with linear regression as the benchmark. Our comparative study demonstrates that models that are capable of capturing the underlying nonlinear interactions between covariates outperform other methods.}


\maketitle

\section{Introduction}\label{sec1:intro}

Over the past decade, there has been a substantial upsurge in the use of social media platforms. This marked growth has been followed by the emergence of new formats of postings such as text, image, video, and hybrid. 
As social media platforms continue to expand their market share in the web service industry, business practitioners are exploring ways to leverage the vast amount of data generated by these platforms. One approach includes assessing the influence of individual users or specific posts. Influencer marketing is an example of such an approach, which is becoming increasingly popular as it has been found to exert a strong effect on social media users' purchasing decisions. 
The number of ``Likes'' on a post is a key indicator of popularity on social media and provides valuable insights into diverse fields, including academia and industry practitioners. The surge of academic interest towards the ``Likes'' is evident in various research fields \citep{ZADEH2022103594, JEON2020120303, mcparlane2014nobody, wu2015analyzing, yang2020named, li2019senti2pop}. 

Predicting the popularity of posts on social media can provide businesses with valuable insights into which type of content resonates with their audience, allowing them to optimize their marketing strategies accordingly. \cite{ketelaar2016success} predict how users on a Dutch social media platform (Hyves) respond to real advertising posts. \cite{sashi2019social} investigate the factors influencing customer behavior in the quick service restaurant (QSR) using Twitter data. These studies exhibit direct and practical implications for 
businesses' successful advertising decisions. As such, it has become crucial for businesses to understand the type of content that is likely to generate more consumer engagement and reach a wider audience. 

However, there are several reasons why predicting the popularity of social media posts can be challenging for researchers or practitioners. First, it is a complex task to adequately incorporate image data into popularity prediction. Modern users on social media platforms communicate more through their image postings rather than through text languages \citep{marwick2015instafame}. In contrast to the numeric and text data collected in social media platforms, such as the number of followers and hashtags, an image file cannot be simply transformed to a small number of numeric or dummy variables. 
To address this issue, some studies use pixel information for images \citep{ding2019social}. However, it remains challenging to discern which aspects of an image contribute to its popularity, as these methods do not offer an explanation of image content. Rather, they directly employ a list of numerical values that are not readily interpretable. 

Second, because social media data has its own unique features, selecting models that would yield superior performance can be a challenging process, especially in popularity prediction. Social media data often exhibits a hierarchical structure, where users upload multiple posts and associate multiple images with each post. As a result, models need to treat individual-level user effects separately in order to account for this complexity. However, previous works have often overlooked the importance of individual user effect in popularity prediction \citep{ de2017predicting,deza2015understanding}.\footnote{Specifically, \cite{deza2015understanding} only use image characteristics without user characteristics for popularity prediction, while \cite{de2017predicting} only use the number of followers as the only feature for capturing user-specific effects.} In addition to such user-specific effect, there may exist complex interactions between covariates both within and across hierarchies, which can affect the popularity prediction of each post. Neglecting these unique characteristics of social media data can significantly impact the accuracy of popularity prediction. For example, Elvis Presley posting a guitar image and Neil Armstrong posting a guitar image could lead to entirely different outcomes. 
Only a few studies have considered this interaction, exceptions being \cite{zaman2014bayesian} as they use a Bayesian hierarchical model and \cite{RISIUS2015824} use a hierarchical linear model to incorporate the hierarchical structure of both user-level and post-level data. However, in \cite{zaman2014bayesian}, sensitivity to the specification of prior distributions and the computational burden of Bayesian approach may be challenging for researchers.

In this paper, we explore various models to predict the popularity of an image-based social media post. Specifically, we compare Linear mixed model, Support Vector Regression, Multi-layer Perceptron, Random Forest, and XGBoost with linear model as the baseline. As covariates, we use commonly used variables in the popularity prediction literature and additionally explore the importance of using social media post image-related variables, such as image labels and representative color, to improve the model fit for popularity prediction.

Our study makes several contributions to the literature. First, we compare the performances of models that can capture the complex structure of social media data. Unlike traditional data, the features of social media data display a hierarchical structure, in the sense that a single user uploads multiple posts, and for a given post, multiple images may exist. In addition, the interactions between covariates both within and across hierarchies, may affect the popularity prediction per post. We find that among the models considered, Random Forest and XGBoost display a better model fit, which is consistent with the fact that these models better reflect the complex structure of the data.

Second, we propose useful methods to extract key image information that improves the model fit for post-popularity prediction in a systematic and replicable manner. Using the Google Cloud Vision API (Google API), we acquire object labels and dominant colors from each image of a single post. Then, we use Seeded Latent Dirichlet Allocation (Seeded-LDA) \citep{seededLDA_1,seededLDA_2} on the attained labels to summarize the image contents. To construct the representative color of a post, we feed the color information obtained from Google API into the Munsell color system.

Third, we propose utilizing interpretable variables, such as topic variables and representative colors.
Experimental results demonstrate that the proposed interpretable variables achieve comparable performance to features constructed by various existing methods, including embedding vectors extracted using deep learning.

Finally, our considered methods are both practical and interpretable, yielding direct implications for real-world applications. Our image processing procedure minimizes the researcher's discretion and systematically automates the process for facile replication. Furthermore, the suggested methods facilitate easier interpretation because the image labels obtained from such processes are in everyday language (food, sky, juice) instead of pixel-level numeric vectors, and the colors obtained from the Munsell color system are consistent with daily color perception by the human eye (purple, yellow, brown), unlike the commonly used numeric Red-Green-Blue color metric.

The remainder of this paper is organized as follows. The literature review is provided in section \ref{sec2:review}. Data collection and variable construction are explained in Section \ref{sec3:data}. Section \ref{sec4:models} presents the models considered for popularity prediction in social media data. The data analysis results are discussed in Section \ref{sec5:results}. Finally, Section \ref{sec6:conclusion} concludes.

\section{Related work}\label{sec2:review}

\subsection{Non-image variables for popularity prediction}

Previous studies on popularity prediction for social media contents document the importance of using non-image information, which includes hashtags, time, and caption. First, hashtags in social media posts serve to summarize the content of the post and increase its exposure to other users. 
\cite{insta_3} and \cite{model_3} focus on the quantitative feature of hashtags by setting the number of hashtags as an input variable. On the other hand, \cite{model_1} utilize the semantics of hashtags. Specifically, they constructed a corpus of words commonly used in hashtags and transformed this corpus into a set of numerical vectors using a word2vec language model. Also, hashtag popularity and the duration of being the top-ranked post in a search result for a specific hashtag have been used to predict popularity for Twitter posts \citep{9311855}. 

Second, temporal data, such as the time of day, day of the week, and month of posting, are commonly utilized in prediction models due to their ease of use \citep{chopra2019comparative, hidayati2017popularity}. Some studies incorporate variables such as whether a post was uploaded on holiday \citep{insta_1}, and the time elapsed between the current and previous posts \citep{figueiredo2013prediction}.
\cite{shulman2016predictability} predict the popularity of posts on social media like Flickr and Twitter using temporal features between the initial sharing of a post and its $i$th ($i=2,\dots,k$) sharing. 
\cite{LI2016310} utilize the time elapsed between image posting and web scraping as a predictor for predicting the popularity of Instagram posts.

Third, caption information is still crucial in understanding the uploaded content, despite the increasing importance of image contents. The most commonly used methods for caption data are based on the idea to check which words within a post are more related to its popularity. 
Generally, analyzing the caption involves examining either the total length of the caption \citep{Flickr_2}, the presence of specific words \citep{naveed2011bad}, and the frequency of words or sentences in the caption \citep{arapakis2014feasibility}. 
Recently, scholars have pointed out that such classic methods do not fully utilize the informational richness of caption data. 
\cite{keneshloo2016} and \cite{SAEED2022116496} use VADER algorithm \citep{VADER} to extract the sentiment of captions as probabilities belonging to either positive, negative, neutral, or compound classes. Some studies utilized complex models such as word2vec \citep{sanjo2017recipe}, Long Short-Term Memory (LSTM) \citep{hessel2017cats}, or pre-trained Bidirectional Encoder Representations from Transformers (BERT) \citep{ding2019social} to summarize the information in the caption.

\subsection{Image variables for popularity prediction}

Various techniques have been proposed to summarize the complex information in images to improve modeling accuracy. One fundamental piece of information is the pixels, which refer to the structural information of the image. Some studies directly use pixel information for prediction. 
For instance, pre-trained deep learning models such as Resnet are utilized to extract deep image features, representing pixel-level information of the image in an abstract manner \citep{ding2019social,9576573}.
\cite{chen2019social} employ Hu moments, a moment-based descriptor, to reduce the dimensionality of Flickr images by calculating the brightness values of pixels in the image. However, such pixel-based covariates are often difficult to interpret since the human eye perceives not individual pixels but rather coherent visual context.

Alternative approaches extract easy-to-interpret semantic features from images using deep learning methods. First, techniques utilizing object detection within images have been developed to identify specific objects. To avoid manual classification of object categories, the output of a deep convolutional neural network was used to identify distinct object categories \citep{huang2017towards, hessel2017cats, chen2016micro, overgoor2017spatio,su2020predicting}. 
Next, some studies have explored using image captioning models to generate textual descriptions of the images. \cite{Hsu} utilize Word2Vec to create representation vectors, while \cite{8622461} perform Latent Dirichlet Allocation (LDA) with 400 topics.

Color information has also been explored as a potential source of information for image processing.
\cite{lv2017multi} divide an image into $k$ sub-images and construct the probabilities of the sub-image pixels containing $m$ focal colors as a basis for a covariate. 
\cite{Pinterest} first define basic colors (e.g., black, blue, brown, etc.) based on the Hue-Saturation-Luminance (HSL) color space and define dominant color of the given image as basic color that cover least 60\% of the image content. 
Similar to our study, \cite{7903630} cluster the Hue-Saturation-Value (HSV) color space into 10 distinct classes based on their coordinates: black, white, blue, cyan, green, yellow, orange, red, magenta, and others. Both studies confirmed that combining color information with other visual features or modality features improved prediction accuracy.

\subsection{Prediction model}

There have been several parametric methods used to predict popularity. For instance, \cite{5616467} used Cox proportional hazards regression models to forecast if comments from two discussion forums, dpreview and myspace, would attract popularity above a certain threshold. The study identified inter-comments time metrics, including the median, mean, and variance, as risk factors compared to the role of smoking in survival analysis. 
\cite{zaman2014bayesian} develop a Bayesian approach for predicting the number of retweets on Twitter posts and the credible intervals for these predictions. Their approach incorporates information on the time of retweets and the local network structure of retweets. They reveal good predictive results, even for tweets posted just a few minutes ago. \cite{FANG2023103329} employ a regression model on panel data from Bilibili video content to examine the influence of social signals on viewership. They established that social support metrics, like cumulative ``Likes'', play a crucial role when viewers select digital content.

Most recent studies tend to adopt machine learning models to improve prediction. Both \cite{Flickr_3} and \cite{chen2019social} aim to predict the number of clicks on Flickr posts. In \cite{Flickr_3}, kernel Support Vector Regression is used, and two distinct datasets (user-mixed, user-wise) are created to investigate the user effect. The results show that controlling for the user effect increased the role of covariates in predicting popularity. As for \cite{chen2019social}, using XGBoost shows that performance is enhanced by integrating user information, non-image features, and image-related features. \cite{purba2020analysis} define popularity as the outsider percent, which means the number of outsiders divided by the number of likers, and predict the popularity of Instagram posts using Random Forest. Through variable importance analysis of Random Forest, they found that the number of hashtags, followers, user tags, and posted hour were the most important factors in predicting popularity, in that order. 
\cite{9243175} treated captions as word sequences and generated a probability vector for adjacent word pairs using a Bayesian network. They then analyzed the relationship between the caption's probability distribution and popularity using Multi-layer Perceptron (MLP) method, although they failed to achieve performance improvements through the MLP.

\section{Data description}\label{sec3:data}

This section describes our data collection process and construction of the variables used in the analysis.

\subsection{Data collection}
\subsubsection{Sampling process}

To collect our sample of users and their post content information, we first selected a focal user. We then limit the next user selection pool to the focal user's following list and select a user, and repeat this process sequentially. The details are described in Appendix \ref{app:usersampling}. The data crawling period is between 6 February, 2022 and 24 March, 2022, with a resulting analysis sample of 40 users, 3,807 posts\footnote{From 4,000 collected posts, those consisting only video contents without any image are excluded.}, and a total of 13,774 images.

\subsubsection{Google Cloud Vision API}

We use the Google Cloud Vision API (Google API) to extract information from the collected images. It uses Google's pre-trained model to extract information from an image, including labels, colors, texts, and places. In many contexts, Google API has served as preprocessing tool, facilitating the extraction of text from images \citep{YU2021103204, API1, API2, API3, API4, API5,NANNE2020156}. We specifically used the `Detect labels' and `Detect image properties' functions in the Google API.

The `Detect labels' function returns labels of objects recognized in an image and their corresponding label score, which is a value representing the estimated accuracy of the detected label. In our analysis, labels with a label score of 0.5 or more were used. Because the default Google API setting returns only up to 10 labels, we adjust the process to return up to 100 labels detected in the picture to obtain as much information as possible.\footnote{Taking Python as an example, features can be adjusted with requests when the image is provided as input to the API. Among them, the `max results' factor can be increased as much as desired. ex) max results = 100.} 

The `Detect image properties' function detects prominent colors in an image. If an image is fed into the Google API, up to 10 dominant colors can be obtained in RGB form--a three-dimensional numerical value denoting the degree of Red, Green, and Blue in a given color--, and their respective pixel fractions, which refer to the fractions occupied by the corresponding colors in the image, are provided.\footnote{Unlike the `Detect labels' function, it should be noted that the maximum result value cannot be adjusted for the number of colors returned. Since the task of extracting `dominant colors' is not aimed at extracting all colors, the maximum results are considered to be limited.} Among the 10 dominant colors returned by the Google API, we choose the color with the highest pixel fraction value and define it as a representative color.

\begin{figure*}[h]
\centering
    \includegraphics[width=0.85\textwidth]{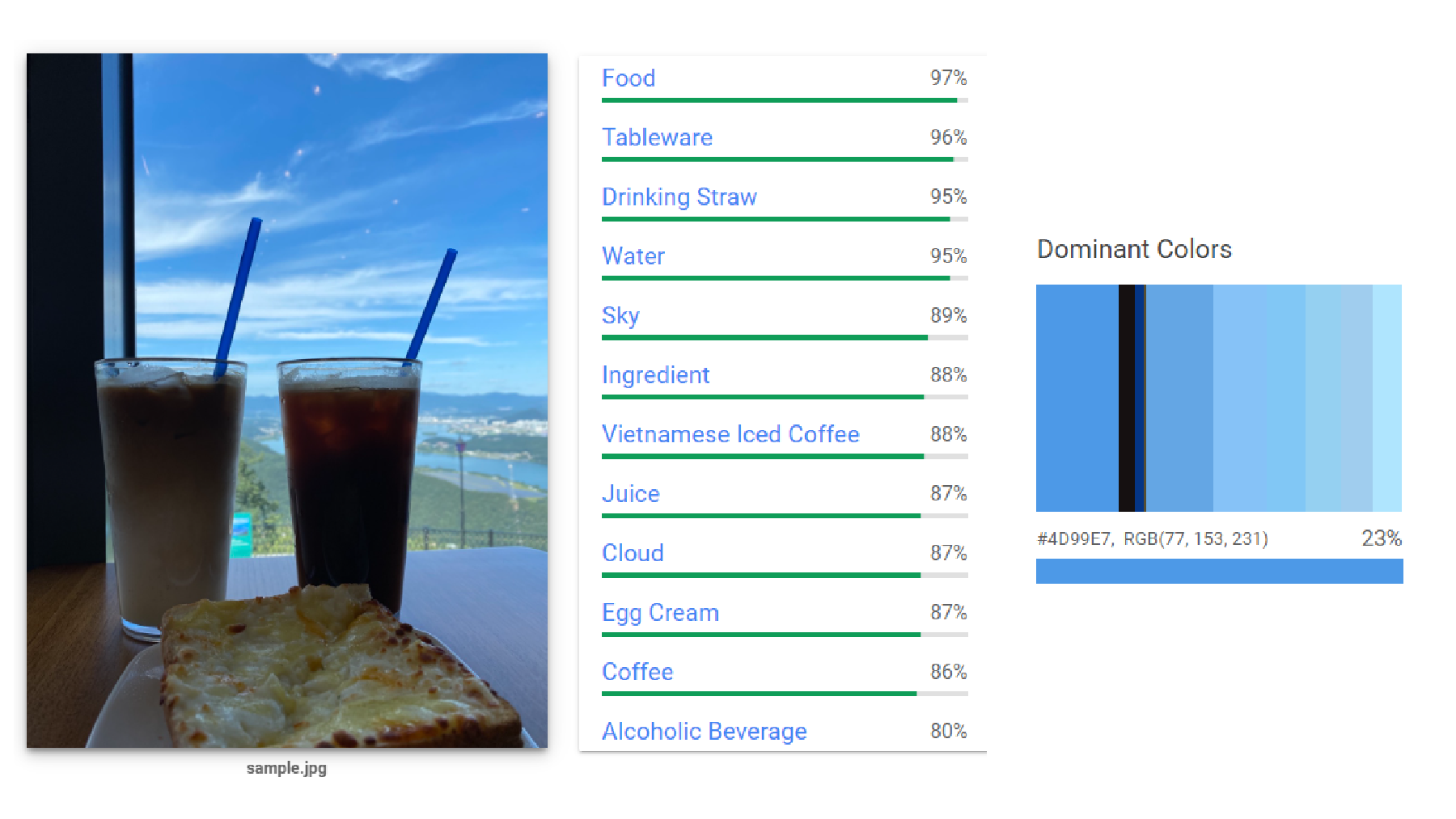}
    \caption{ Google Cloud Vision API output of a sample image. From left to right: the sample image, returned labels/label scores, and dominant colors/representative color}
    \label{fig:API}
\end{figure*} 

Figure \ref{fig:API} shows an example of how Google API is used. It shows the results obtained by applying the two aforementioned functions to a sample image. An image taken by the author of food and iced coffee at a caf\'e is used as the input in the leftmost column. The second column contains the returned labels, such as Food, Tableware, and Sky, and the associated label scores (in percentages). In the rightmost column, the dominant colors and their pixel fractions are displayed, which may refer to the color of the sky or drink, depending on what color is emphasized in the image. We define a representative color of this image as the dominant color with the highest pixel fraction, which is the color with RGB (77, 153, 231) in this case.

\subsection{Variable construction}

\begin{table*}[!htbp]
\caption{Variable Description}
\label{tab:variables}
\small
\centering
\begin{tabular}{llll}
\toprule
Variable & Description \\
\cmidrule(l{0pt}){1-1} \cmidrule(l){2-2}
Likes & Log (number of ``Likes''/ Time difference) \\
User& User-specific indicator  \\
N. of Image & Number of images per post  \\
N. of Reels & Number of short videos per post  \\
Public & Indicator for whether the number of ``Likes'' is private or not \\
Time difference & The difference between the time of crawling and posting \\
Period & The difference between the upload time of the current post and the previous post \\
Weekdays & Indicator for the posting is made on weekdays \\ 
Hour & Indicators for posting time in 3-hour intervals  
\\
Holiday & Indicator for legal holidays \\
Season & Indicators for the season of posting - Spring, Summer, Fall, Winter \\
Tagged place & Indicator for whether a place is tagged on a post \\
N. of Tagged id & Number of tagged users \\
N. of Hashtag & Number of hashtags \\
Caption topic & Indicators for each of the caption topics - Event, Health, Beauty, Fashion, Daily \\
Image label topic & Indicators for each of the Image label topics - Food, Body, Beauty, Fashion, Daily \\
Representative color & Indicators for the representative color category - R (red), YR (yellow red), \\ 
& Y (yellow), GY (green yellow), G (green), BG (blue green), B (blue), \\
& PB (purple blue), P (purple), RP (red purple) \\
\bottomrule
\end{tabular}
\end{table*}

After the data collection phase, we construct variables to be used in the prediction model, which capture essential information from the raw data. A detailed description of the variables used in this study is provided in Table \ref{tab:variables}.
Our approach focuses on constructing interpretable variables, in the sense that the variables align with human perception thereby yielding interpretable prediction results. For instance, when the topic variable of a given image is `Trekking', the post popularity may be higher when its associated `color' variable is `Green', whereas a topic variable corresponding to `Swimming' may exhibit distinct association with color `Blue' in predicting ``Likes''. Such improvements in interpretability could enhance the understanding of data analysis results.

\begin{figure}[!h]
  \centering
  \begin{tikzpicture}[node distance=0.4cm]
    \node (img1) {\includegraphics[scale=0.25]{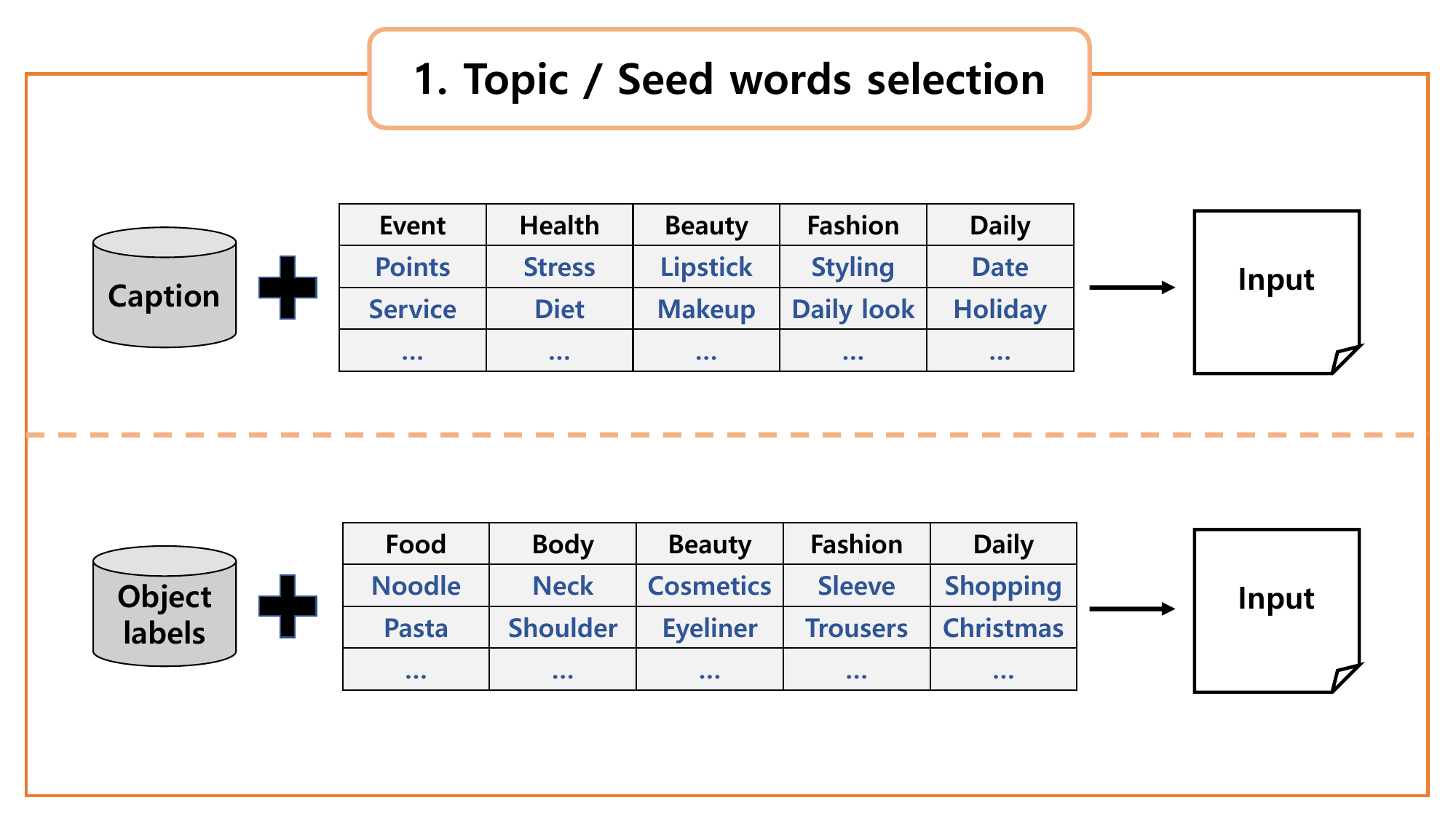}};
    \node[below=of img1] (img2) {\includegraphics[scale=0.25]{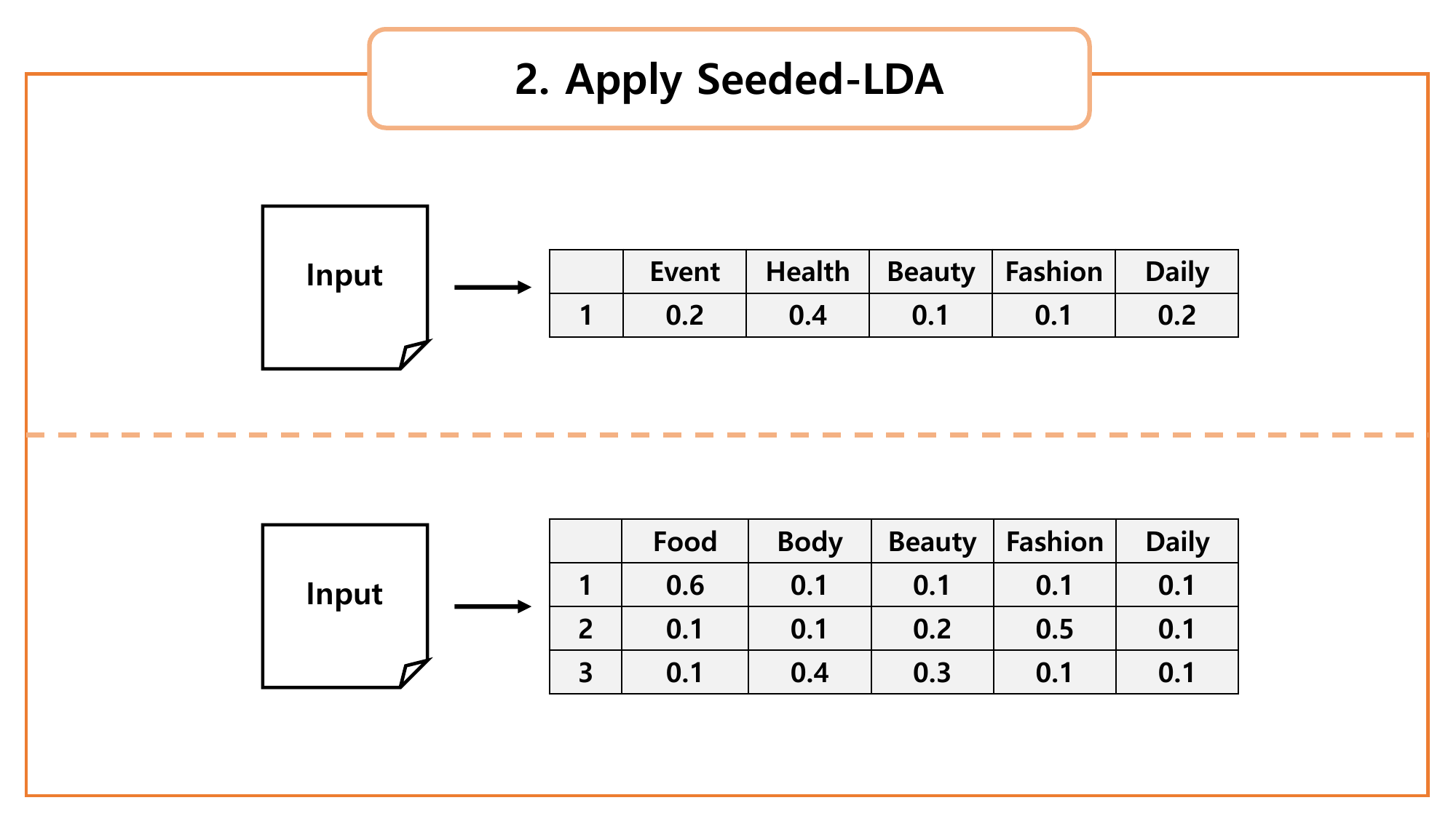}};
    \node[below=of img2] (img3) {\includegraphics[scale=0.25]{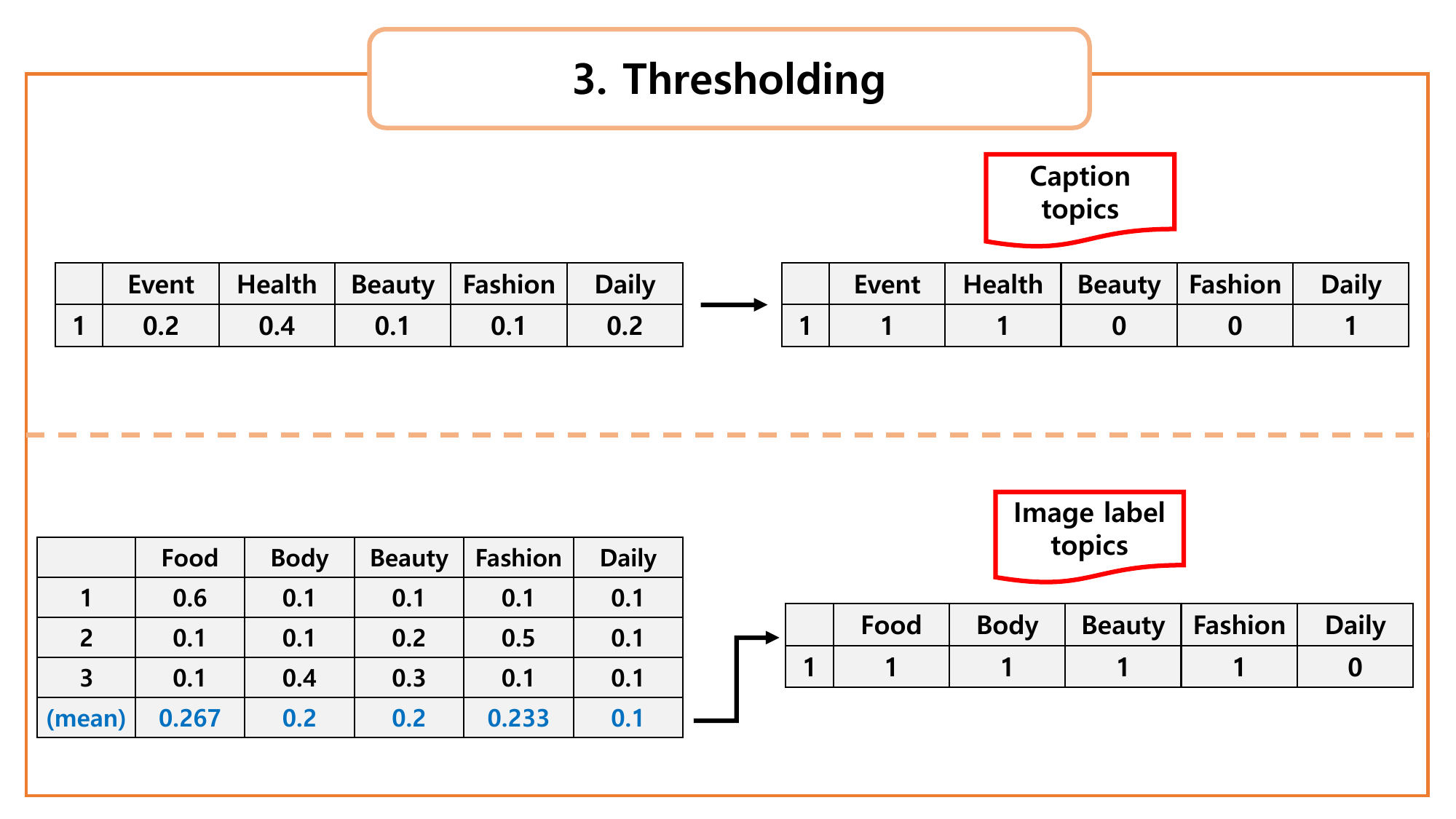}};

    \draw[-stealth] (img1) -- (img2);
    \draw[-stealth] (img2) -- (img3);
  \end{tikzpicture}
  \caption{Topic variable construction: This figure provides an example of a case where a single post contains three images}
  \label{fig:topic_var}
\end{figure}

\subsubsection{Topic variable construction for captions and image labels \label{subsubsec:topic}}

Information on the content of a social media post, namely captions and image labels, consists of a large collection of words. As the scale of the corpora reaches over a thousand words, dimension reduction is recommended to avoid the issue of overfitting and reduce the computational burden. In our analysis, we employ a topic modeling method, namely, Seeded Latent Dirichlet Allocation (Seeded-LDA) \citep{seededLDA_1,seededLDA_2} for dimension reduction. 
Seeded-LDA generates distinct topics for image labels and post captions. Using a document's identified topics as covariates enhances our ability to comprehend the individual influence of each topic on the response variable.
Also, the explicit availability of the word distribution for each topic allows for a more detailed analysis of the input topic covariate, leading to a higher level of interpretability of the results.

The process of constructing topic variables is illustrated in Figure \ref{fig:topic_var}.
In Seeded-LDA, topics and seed words can be manually predefined according to the researcher's prior knowledge of document features \citep{seededLDA_1, seededLDA_2}. We assume that captions and images in social media data can be characterized by dominant topics.
Specifically, we choose five topics for both the captions and image labels by manually exploring the analysis sample. Topics for captions include `Event', `Health', `Beauty', `Fashion', `Daily', and the topics for image labels are `Food', `Body', `Beauty', `Fashion', `Daily'. 

Next, we choose seed words for each topic based on the frequency of words across all captions/image labels. Given that even a small set of seed words improves the prediction performance \citep{seededLDA_1}, several top--frequency words are selected as seed words. Specifically, for each caption topic, the top 20 words out of 10,810 words were selected; for each Image label topic, the top six words out of 4,542 words were selected as seed words.\footnote{As the corpora of the caption are derived from natural language, it is much larger than that of image labels. Hence, we select fewer seed words for Image label topics than caption topics.} The seed words for each topic and hyperparameters used are illustrated in Appendix \ref{app:slda_param}. 

The Seeded-LDA provides posterior probabilities for caption topics and Image label topics as outputs. Regarding Image label topics, because a single post may consist of multiple images, we construct a single post-level variable that reflects information across images for a given post for post-level analysis using the mean of posterior probabilities for each Image label topic. As there is a single caption for a single post, the posterior probability for the caption was used as is. Finally, for better interpretability, we construct indicator variables for each topic that take a value of one if its associated probability is greater than a given threshold.\footnote{
We employ binarization as a denoising technique to address the inherent noise present in topic probability vectors achieved from Seeded-LDA. 
Since LDA typically assigns non-zero, albeit potentially small, probability values to all topics for a given document, the binary approach aims to strengthen the document's association with its most relevant topics and reduce the influence of less relevant ones. 
This approach has been shown to improve results in previous research \citep{hernandez2017deceptive}, and we observed similar findings in our own experiments.} Since there are a total of five topics, with the associated probabilities summing to one, the value of 0.2 (1/number of topics) is used as the threshold.\footnote{
As Seeded-LDA can be sensitive to the selection of several pre-determined topics and seed word sets, we perform sensitivity checks in Appendix \ref{app:slda_topic} and \ref{app:slda_seed}. Our results exhibited indifference to different topic diversities and seed word sets.}

\subsubsection{Image color variable \label{subsubsec:img_color}}

We consider image colors as meaningful information for predicting image-based social media post popularity because colors can evoke emotional responses and visually attract users. Research suggests that certain colors may elicit specific emotions or grab more attention, potentially influencing engagement with the content (e.g., \cite{machajdik2010affective, wei2006image}). 

Specifically, we define `Representative color' of an image because individuals tend to recognize the dominant colors in an image, rather than perceiving all the colors. In other words, the overall color tone may be more informative for popularity prediction, rather than the full set of colors displayed in the image. Dominant colors are often utilized in social media image analysis research to investigate the link between color and user engagement (e.g., \cite{aryafar2014exploring, wang2023pictorial}), including studies that use dominant color for image popularity prediction \citep{7903630,Pinterest}. In line with such studies, we incorporate representative color covariates into our analysis to account for the influence of color on prediction accuracy.

In our analysis, for a given image, we first obtain the RGB value of its representative color through the Google API. Then, we convert the RGB value of its representative color into one of the ten basic colors of the Munsell color system\footnote{The function \texttt{col2Munsell} in package \texttt{aqp} in \texttt{R} is used. The conversion method can be referred to in \href{https://www.rit.edu/science/munsell-color-science-lab-educational-resources}{The Rochester Institute of Technology (RIT) website}.}\citep{munsell_color}, which consists of red (R), yellow-red (YR), yellow (Y), green-yellow (GY), green (G), blue-green (BG), blue (B), purple-blue (PB), purple (P), and red-purple (RP). At the post-level, the `Representative color' variable is coded as a binary dummy variable. For example, if a post consists of four images and the images have representative colors of Y, Y, BG, and B, the color category variable is coded as $(0,0,1,0,0,1,1,0,0,0)$ where the variables are in the order of R, YR, Y, GY, G, BG, B, PB, P, and RP.

\begin{figure*}[htbp]
\subfigure[Number of ``Likes'']{
\centering
\includegraphics[width=.3\textwidth]{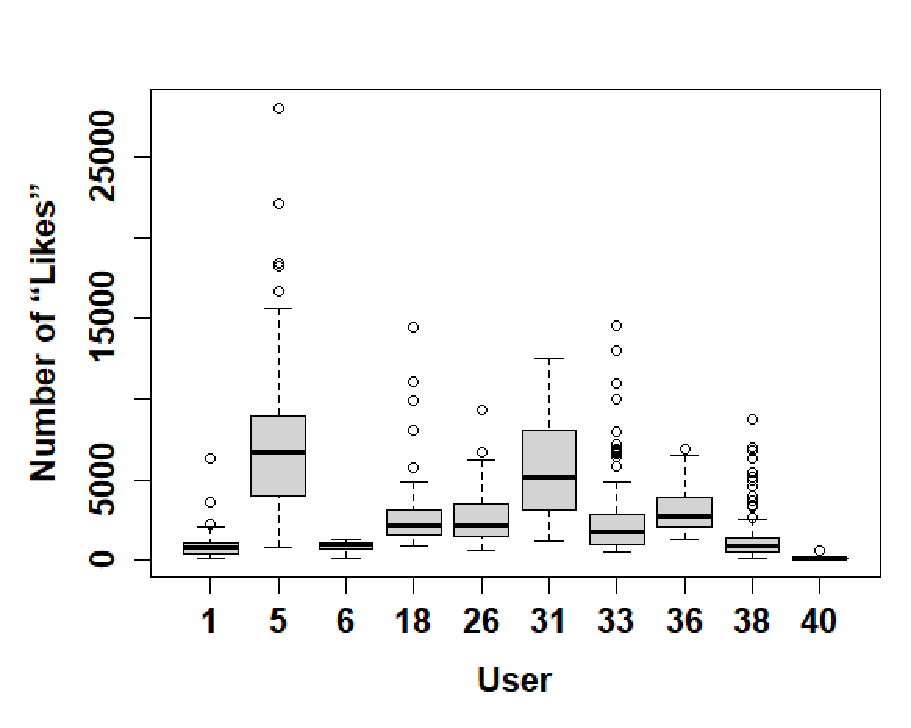}
}
\subfigure[Number of ``Likes''/ Time difference]{
\centering
\includegraphics[width=.3\textwidth]{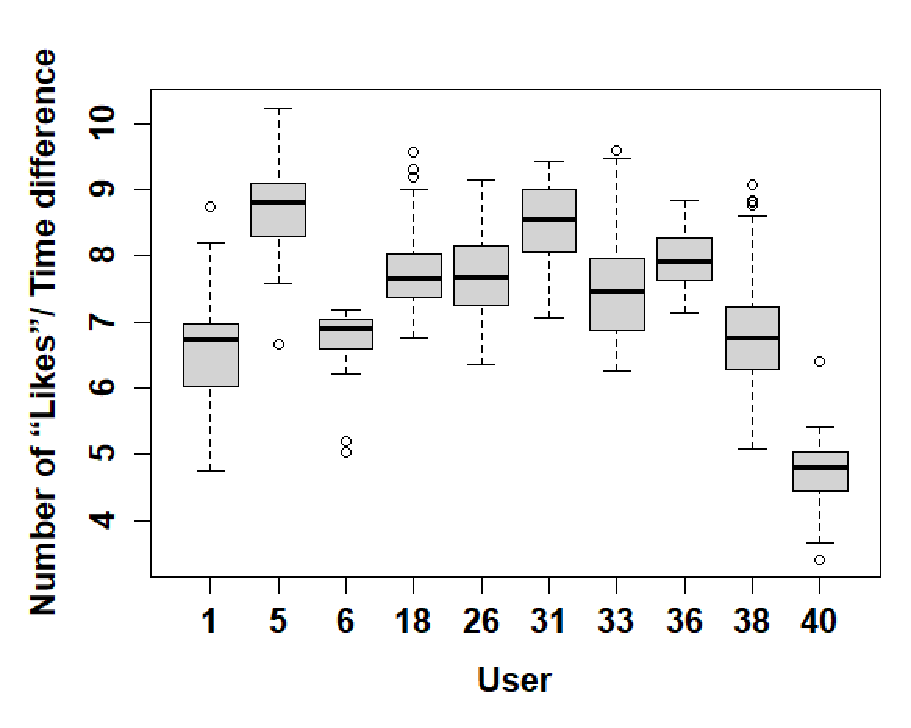}
}
\subfigure[Log (number of ``Likes''/ Time difference)]{
\centering
\includegraphics[width=.3\textwidth]{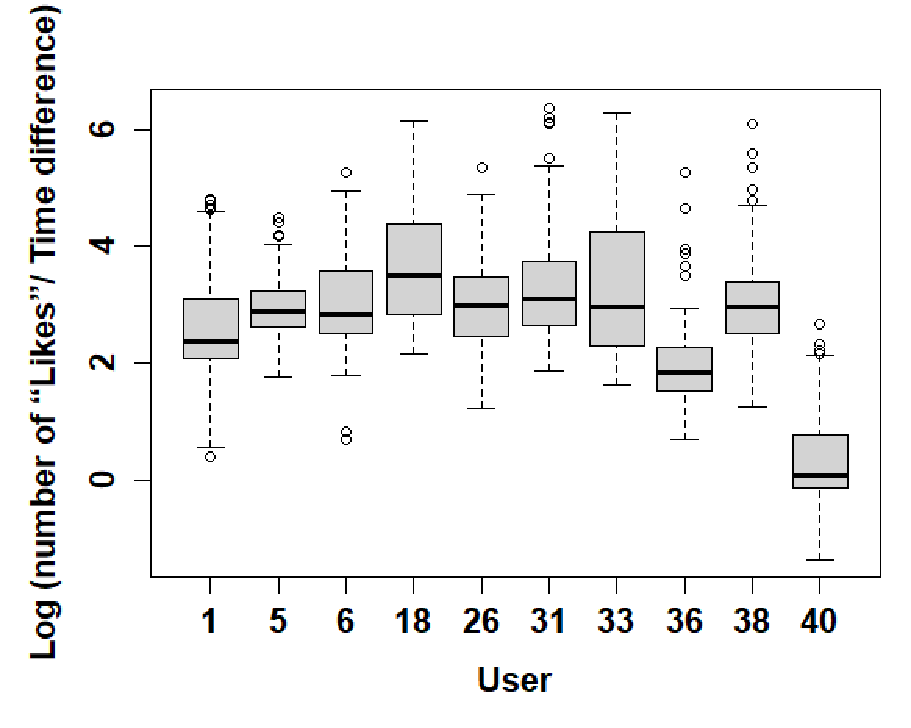}
}
\caption{ Panel (a) shows boxplots of the number of ``Likes'' for all users. Panel (b) is the response variable in (a) scaled by `Time Difference'. Panel (c) log transforms the response variable in (b)}
\label{fig:like} 
\end{figure*}

\subsubsection{The other variables}

Below, we provide an explanation for the remaining explanatory variables and the response variable.

\begin{itemize}
   \item The response variable: ``Likes''
   \medskip
   
       We log-transform the time-adjusted number of ``Likes''. Figure \ref{fig:like} displays the effect of the time-adjustment and the log transformation. The first column is the raw number of ``Likes'', which shows wide variation across user-wise ``Likes'' distribution. In the second column, we account for the `Time Difference' (the difference between the crawl time and the upload time) by scaling `Like/(Time Difference+$c$)', which moderates the variation, consistent with previous studies \citep{Flickr_3, Flickr_1}.\footnote{In our analysis, $c=5$ was chosen as it empirically shows good results in reducing the users' within variance.} This is to correct the fact that if ``Likes'' are collected as soon as the post is uploaded, the number would be relatively small regardless of the potential popularity of the post. Then, log transformation to correct for right-skewness was applied in the third column. In our analysis, we used `log(Like/(Time Difference+$c$))' as a response variable.

   \medskip
   
    \item Time-related variables: `Weekdays', `Hour', `Season', `Holiday', `Period', and `Time difference'
   \medskip    
   
    Basic time-related variables such as a weekdays dummy, time of the day, season dummies, and a dummy for national holidays were used, consistent with previous studies \citep{Flickr_2, model_1, insta_1}. The `Weekdays' is a variable indicating whether the post is made on a weekday. The `Hour' variable denotes time of the day in a 3 hour interval. The seasonality is controlled by using dummies. The `Holiday' indicates whether the post's upload day is a national holiday or not. 
    
   \medskip
   
    To capture the frequency of posts for a given user, a variable `Period' is constructed. It refers to upload time interval between posts, calculated by the time difference between current and previous posts. The frequency of users' posts may either induce more interest or fatigue the audience.
    We additionally include a variable `Time Difference' to capture the difference between the time of crawling and that of post upload. This is to control for the fact that due to intermittent crawling, the number of ``Likes'' may differ depending on the time interval between posting and crawling. 

   \medskip
   
    \item The user-specific effect: `User'
   \medskip    
   
    In previous studies, various user characteristics expected to affect popularity were used as covariates, such as user history (mean and standard deviation of popularity) and user profile \citep{Flickr_3}. Instead, we include user fixed effects, or a dummy for each user, which captures both observable and unobservable characteristics of user-specific effects, such as the number of followers, total number of posts, gender, and age of the user. 

   \medskip
   
    \item Content related variables: `N. of Image' and `N. of Reels'
   \medskip    
   
     The types of content that can be uploaded on social media platforms include image, reels, and IGTV.\footnote{`Reels' means short videos of less than 60 seconds, and `IGTV' means long videos of less than 60 minutes.} Because both reels and IGTV are in a video format, they could not be used as inputs for Google API and were hence excluded from the analysis.\footnote{Video's `thumbnail' could have been used, but since there is a large difference in the amount of information regarding the post in a single thumbnail and a video, we excluded the video contents from the analysis. }
     Instead, we used the number of images and reels included in a given post as variables, as in \cite{insta_3}.\footnote{Note that on social media platforms, images and reels can be uploaded at the same time and are limited to a maximum of 10. However, only one video can be uploaded when uploading IGTV.}
  
   \medskip
     
    \item Tag related variables: `Tagged place', `N. of Tagged id', and `N. of Hashtag'
    \medskip   
    
    The tag information that can be obtained in a single post includes a location, other users' id, and hashtags. Tag information has been used in most previous studies \citep{Flickr_2, insta_3, Flickr_3, model_1, model_3, model_2, insta_1}.
    We consider variables that indicate whether a post has a tagged place, the number of users tagged through the image of the post, and the number of hashtags used in the caption of the post. Using multiple tag information when uploading a post increases the likelihood of more users encountering the user's post.

   \medskip
   
    \item The disclosure of the number of ``Likes'': `Public'
    \medskip
    
    We include an indicator for whether other users could see the number of ``Likes'' on each post. Recently, social media platforms have added a feature for users to choose whether they want to disclose the number of ``Likes'' of posts in the feed to other users. This function was introduced to relieve users of the burden of others' reactions. Consequently, users could make the number of ``Likes'' private if they felt burdened by the low ``Likes'' of their posts.\footnote{In this case, only the list of users who liked the post is visible. Therefore, we crawled the list of users who `Liked' a post and counted the number of users to calculate the raw number of ``Likes''.}

\end{itemize}

\subsubsection{Preliminary analysis} 

To explore whether the variables used in our analysis are useful in ``Likes'' prediction, we perform a preliminary analysis using linear regression. We use three sets of variables, where the first set of variables includes variables commonly used in previous studies.\footnote{Public, N. of Image, N. of Reels, Tagged place, N. of Tagged id, N. of Hashtag, Holiday, Season, Weekdays, Hour, Period, and Time difference} The second set of variables include user-specific dummies. The third set of variables is constructed using Google API in Sections \ref{subsubsec:topic} and \ref{subsubsec:img_color}.

\begin{table*}[h]
\caption{Preliminary analysis result}
\label{tab:Prelim_analy}
\centering
\small
    \begin{tabular}{cccc}
    \toprule
    & Model 1 & Model 2  &  Model 3  \\
    \cmidrule(l{0pt}){1-1} \cmidrule(l){2-4}  
    Common variables & O & O & O    \\
    User & X & O  & O    \\
    Caption topics \& Image label topics \& Colors & X & X &O \\
    \cmidrule(l{0pt}){1-1} \cmidrule(l){2-4} 
    $R^2$ & 0.3031 & 0.7586 & 0.7849  \\
    Adj. $R^2$ & 0.2987 & 0.7541 & 0.7795  \\
    \bottomrule
    \end{tabular}
\end{table*}

We perform a linear regression by sequentially including the aforementioned three sets of variables, as shown in Table \ref{tab:Prelim_analy}. Compared to Model 1, which only includes common variables as covariates, Model 2 displays large improvements in terms of adjusted $R^2$, once unobserved user characteristics are controlled for by including `user dummies'. This result reflects that our data display hierarchy in the sense that given a user, there are multiple posts belonging to him/her, and consequently, ``Likes'' of posts are highly dependent on the user.

\begin{figure*}[t]
    \begin{center}
    \caption{ Residuals of randomly chosen 10 posts of 20 users of Model 1 and 2. Vertical gray bars mark a partition between distinct users}
    \label{fig:pre_model}
    \subfigure[Residuals of Model 1]{
    \includegraphics[width=0.39\textwidth]{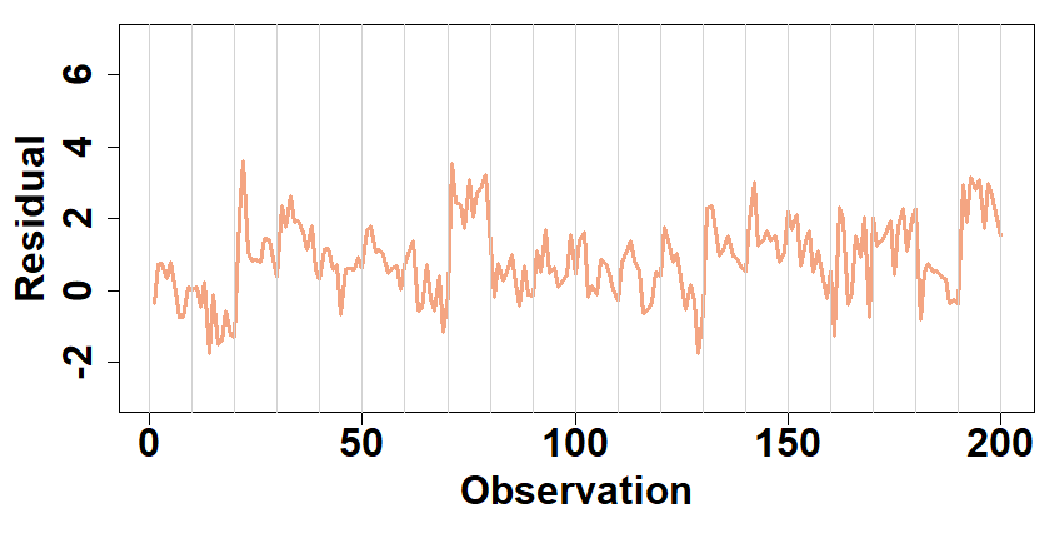}}
    \quad
    \subfigure[Residuals of Model 2]{
    \includegraphics[width=0.39\textwidth]{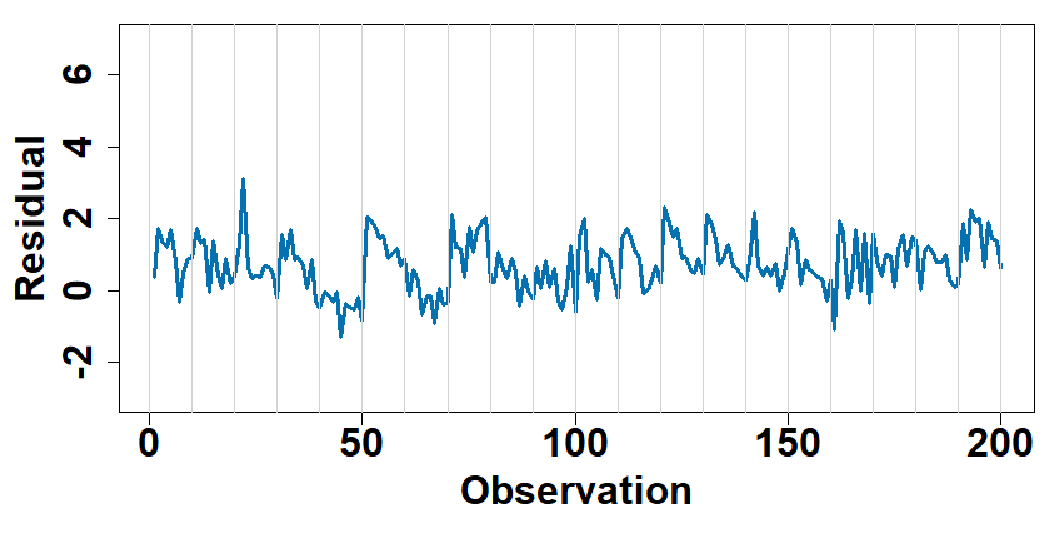}}    
    \end{center}    
\end{figure*}

We also observe from Figure \ref{fig:pre_model} the importance of considering such a hierarchy in predicting ``Likes'' for a post. Panels A and B of Figure \ref{fig:pre_model} display the residual plots of 10 posts for each of the 20 users for Models 1 and 2, respectively. By comparing Panels A and B, we find that the range of the residuals is reduced once the user dummies are included in Model 2. 

Model 3 then adds the final set of variables, which are those constructed using Google API. This inclusion further enhances the model fit, supporting that the constructed variables successfully capture meaningful information from the user-posted content, both in texts and in images. 

From this preliminary analysis, we show that using information from caption and image features together yields better performance in modeling ``Likes''. 

\section{Considered models}\label{sec4:models}

We use linear model as a benchmark and explore models that may better reflect the underlying latent data structure of social media data, such as hierarchy and non-linear interactions between covariates. Specifically, we consider Linear Mixed Model (LMM), Support Vector Regression (SVR), Multi-layer Perceptron (MLP), Random Forest (RF), and XGBoost (XGB) to predict ``Likes''. 
We select comparison prediction methods based on a well-balanced combination of predictive performance, interpretability, and computational efficiency.

\subsection{Linear Mixed Model (LMM)}
 
When repeated users are present in the dataset, the LMM accounts for effects that are specific to a given user separately from those that do not vary with users \citep{LMM}. 
Denoting the number of observations under the $j$th user by $n_j$, the response variable vector and the covariate matrix of the $j$th user by an $n_j$-variate vector $\mathbf{y}_j$ and an $n_j \times p$ matrix $\mathbf{X}_j$ respectively, the LMM models the response of the $j$th user as follows:
\begin{align*}
     \mathbf{y}_j = \mathbf{X}_j\beta + \mathbf{Z}_j \mathbf{u}_j + \mathbf{\epsilon}_j
\end{align*}
where the $k$-variate vector $\mathbf{u}_j$ and the $n_j \times k$ matrix $\mathbf{Z}_j$ are the random effects and their associated design matrix for the $j$th user, respectively.
Additionally, $\epsilon_j$ represents the error terms with $\epsilon_j \sim N(\mathbf{0}, \sigma^2 \mathbf{I})$. In this model, the user-specific effects are addressed by $\mathbf{u}_j$, which facilitates distinct models for each user. The random effect $\mathbf{u}_j$ is assumed to be a random vector following a multivariate normal distribution with $\mathbf{u}_j \sim \mathbf{N}(\mathbf{0}, D)$ such that each model attains variation from the fixed effects $\beta$.

\subsection{Support Vector Regression (SVR)}

SVR finds a linear function of the features that retains most of the residuals within a given margin, allowing for limited amount of exceptions. Because of this feature that it controls the magnitude of each of the residuals, SVR is robust to outliers \citep{svr} and has shown competitive performance in popularity predictions, e.g., see \cite{insta_4} and \cite{7903630}.
Denoting the $i$th response variable and its associated $p$-variate covariate vector by $y_{i}$ and $\mathbf{x}_{i}$ respectively, SVR fits the responses by
a linear function of the covariates under several constraints as follows:
\begin{align*}
    \hat{y} &= \langle \hat{\beta}_1, \mathbf{x} \rangle + \hat{\beta}_0, \\
    (\hat{\beta}_0, \hat{\beta}_{1}) &= \underset{(\beta_0, \beta_{1})  \in \mathbb{R}^{p}}{\arg \min} \frac{1}{2} ||\beta_{1}||_2^2 + C \sum_{i}  (\xi_{i} + \xi_{i}^{*}) \\
\text{subject to} \; &\left(\langle \beta_1,  \mathbf{x}_{i} \rangle + \beta_0\right) - y_{i} \leq \epsilon + \xi_{i},  \\
& y_{i}-\left(\langle \beta_1,  \mathbf{x}_{i} \rangle + \beta_0\right)  \leq \epsilon + \xi_{i}^{*},\\
& \text{and} \quad \xi_{i}, \ \xi_{i}^{*} \geq 0
\end{align*}
where $C>0$ and $\epsilon>0$ are the predetermined values.
The constraints of the SVR promote most of the residuals to lie within the desired range of $(-\epsilon, \epsilon)$ with a few exceptions. When either $\xi_{i}$ (or $\xi_{i}^{*}$) is strictly positive, its magnitude of corresponding residual exceeds $\epsilon$ by $\xi_{i}$ (or $\xi_{i}^{*}$), and the residual falls outside $(-\epsilon, \epsilon)$.
The hyperparameter $C$ controls the total amount of exceptions by regularizing the magnitude of $\sum_{i} (\xi_{i}+\xi^{*}_{i})$.

\subsection{Multi-layer Perceptron (MLP)}

MLP is a feedforward neural network consisting of multiple layers with neurons in adjacent layers being fully interconnected. Specifically, the $i$th neuron on the $(l-1)$th layer is connected to all the neurons in the $l$th layer through weights as follows:
\begin{align*}
    a_j^{(l)} = \sigma\left(\sum_i W_{ij}^{(l)} a_i^{(l-1)} + b_j^{(l)}\right),
\end{align*}
where $a_j^{(l)}$ denotes activation the of the $j$th neuron on the $l$th layer,
$W_{ij}^{(l)}$ denotes the weight between the $i$th neuron in layer $l-1$ and $j$th node in layer $l$, 
$b_j^{(l)}$ denotes the bias term for the $j$th neuron in layer $l$, and $\sigma(\cdot)$ denotes an activation function. In MLP, all the activations from the $(l-1)$th layer serve as inputs to a neuron in the subsequent $(l+1)$th layer. The input layer receives the $p$ features of data $\mathbf{x}$ as activations for its neurons, and the activations in the output layer represent the prediction result. Employing a nonlinear activation function in place of $\sigma$ facilitates MLP to learn complex data structure through the sequential processing of information across stacked layers.

\subsection{Random Forest (RF)}

RF is a tree-based method that combines multiple decision trees using bagging \citep{rf}.
In performing RF, multiple bootstrap samples are generated and a decision tree is independently fitted to each sample. The RF then outputs an average of the fitted results across the generated trees. Specifically,
denoting the number of bootstrap samples by $B$ and the tree constructed from the $b$th sample by $T_b$, the RF estimate at covariate $\mathbf{x}$ is as follows:
\begin{align*}
    \hat{y} = \frac{1}{B} \sum_{b=1}^B T_b(\mathbf{x}).
\end{align*}
When constructing a tree in the procedure, not all but a randomly chosen subset of covariates is used at each candidate split, so that correlations among the trees can be alleviated. Using this line of procedures, the RF attains more stable results than a single tree while maintaining the advantages of tree structures, which can capture the non-additive effects of covariates in a data-adaptive manner.

\subsection{XGBoost (XGB) \label{sub:xgb}}

XGB is a branch of the gradient boosting tree method that ensembles multiple decision trees in a sequential manner \citep{XGBoost}. While RF simultaneously combines decision trees, the gradient boosting method successively learns a tree that improves the previous fit and then updates the fit by adding a new tree. At each tree learning step, XGB includes additional penalty terms that regularize the fit of the tree, thus avoiding the risk of overfitting.
Denoting the tree at the $t$th step by $\hat{f}_t$, the fitted value of the $i$th observation after the $t$th step by $\hat{y}_i^{(t)}$, and the loss function by $l(\cdot, \cdot)$, the idea of the XGB is to find $\hat{f}_t$ and update the fit to $\hat{y}_i^{(t)} = \hat{y}_i^{(t-1)} + \hat{f}_t(\mathbf{x}_i)$ at the $t$th step as follows:
\begin{align*}
    \hat{f}_t = \arg \min_{f_t} & \ \sum_i l \left( y_i, \hat{y}_i^{(t-1)} + f_t (\mathbf{x}_i)\right) + \Omega(f_t) 
\end{align*}
where $\Omega(f_t)$ is the regularization applied on the $t$th tree. Specifically, $\Omega(f) = \gamma T + \frac{1}{2} \lambda \vert\vert w \vert\vert^2$ for $\gamma,\ \lambda > 0$ where $T$ denotes the number of leaves, and $w$ is the vector of the leaf weights of the given tree $f$. In performing XGB, approximated optimization is utilized to reduce computational costs.

\section{Data analysis}\label{sec5:results}

\subsection{Tuning parameter selection via hierarchical cross validation}

Among the methods considered in this section, MLP, SVR, RF, and XGBoost require tuning parameter selection. To select the tuning parameters, we use hierarchical cross validation, that accounts for the data structure. 
The dataset was divided into six folds, ensuring that each fold maintains an approximately equal proportion of all 40 users. One fold was used as the test set for evaluation, while the remaining five folds were used for training. Within the training set, we employed 5-fold cross-validation for parameter tuning. Each parameter combination was evaluated five times, and the combination resulting in the lowest average error was chosen. The model paired with the chosen parameters was then applied to the test set to evaluate the test error. We use the cross-validation root mean squared error (RMSE) for parameter selections.
The tuning parameters of each method and their selected values are presented in Appendix \ref{Appendix:tune_result}. All parameters are selected by a grid search.

Our data display hierarchy in the sense that for a given user, there are multiple posts belonging to him/her and consequently, ``Likes'' of posts are highly dependent on the user. To properly address this user-post hierarchical structure in ``Likes'' prediction, a similar distribution of user profiles across all $k$ number of folds is important for estimating the model performance using cross-validation. Without considering the hierarchical structure, certain folds may be imbalanced and not representative of the analysis sample, which may result in biased test errors.

\subsection{Results}\label{sec5:main_res}

\begin{table*}[h]
\caption{Variables included in each group in Section \ref{sec5:main_res}}
\centering
\small
\label{tab:var_type}
    \begin{tabular}{ll}
    \toprule
    Group &  Variables \\
    \cmidrule(l{0pt}){1-1} \cmidrule(l){2-2}
    Common & Likes, User, Time Difference \\
    Non-image & N. of Image, N. of Reels, Public, Weekdays, Hour, Holiday, Season, \\
    & Period, Tagged place, N. of Tagged id, N. of Hashtag, Caption topic \\
    Image & Image label topic, Representative color \\
    \bottomrule
    \end{tabular}
\end{table*}

We classify the variables into three groups: common, non-image, and image variables. The variables included in the common, non-image, and image groups are listed in Table \ref{tab:var_type}. We compare test errors from the following four prediction settings. First, 
\textit{\textbf{Setting-Common}} includes the common group only, serving as a baseline. Next, \textit{\textbf{Setting-NonImage}} includes the variables in the non-image and the common groups. \textit{\textbf{Setting-Image}} includes the variables in the image and the common groups. Finally, \textit{\textbf{Setting-All}} includes all covariates, namely common, non-image, and image variables. The prediction results are presented in Table \ref{tab:analy_CV}.

\begin{table*}[h]
\caption{Test errors based on hierarchical 5-fold CV result. (1) Column-wise minimum error is starred (2) row-wise minimum error is underlined} 
\label{tab:analy_CV}
\centering
\small
\belowrulesep = 0pt
\aboverulesep = 0pt
\setlength{\tabcolsep}{2pt}
\subtable[RMSE]{ 
    \begin{tabular}{cc|cccc} 
    \toprule
    \multirow{2}{*}{} & \multicolumn{1}{c}{} & \multicolumn{4}{c}{Settings} \\
    & \multicolumn{1}{c}{} & All & Non-image & Image & Common \\ 
    \cmidrule(l{0pt}){3-6}
    \multirow{6}{*}{Methods} & LM  & \underline{0.7282} & 0.7559 & 0.8094 & 0.8365 \\ 
    & LMM & \underline{0.6862} & 0.7212 & 0.7842 & 0.8365 \\
    & SVR & \underline{0.6257} & 0.6620 & 0.6604 & 0.6584 \\
    & MLP & \underline{0.5678} & 0.6759 & 0.6406 & 0.6939 \\
    & RF  & \underline{0.5721} & 0.6075 & 0.5800 & 0.6420 \\
    & XGB & 
    \hspace{0.6mm} \underline{0.4941}$^{*}$ & 
    \hspace{0.6mm} 0.5625$^{*}$  & 
    \hspace{0.6mm} 0.5269$^{*}$  & 
    \hspace{0.6mm} 0.6378$^{*}$\\
    \bottomrule
    \end{tabular}
}
\hspace{5mm}
\subtable[MAE]{
    \begin{tabular}{cc|cccc}
    \toprule
    & \multicolumn{1}{c}{} & \multicolumn{4}{c}{Settings} \\
    & \multicolumn{1}{c}{} & All & Non-image & Image & Common \\
    \cmidrule(l{0pt}){3-6}
    \multirow{6}{*}{Methods} & LM  & 0.5569 & \underline{0.5564} & 0.6129 & 0.6258 \\
    & LMM & \underline{0.5218} & 0.5325 & 0.5874 & 0.6257\\
    & SVR & \underline{0.4623} & 0.4741 & 0.4946 & 0.4784\\
    & MLP & \underline{0.4478} & 0.4961 & 0.4797 & 0.4984 \\
    & RF  & \underline{0.4132} & 0.4341 & 0.4166 & \hspace{0.6mm} 0.4588$^{*}$ \\
    & XGB & 
    \hspace{0.6mm} \underline{0.3541}$^{*}$ & 
    \hspace{0.6mm} 0.3978$^{*}$ & 
    \hspace{0.6mm} 0.3879$^{*}$ & 
    0.4657\\
    \bottomrule
    \end{tabular}
    \hfill
}
\end{table*}

Overall, across all methods, \textit{\textbf{Setting-All}}, which includes both non-image and image variables, returns the smallest test error compared to other settings. Also, consistently across all methods, \textit{\textbf{Setting-Common}} exhibits higher error, as measured by both RMSE and MAE, compared to \textit{\textbf{Setting-All}}. In terms of RMSE, \textit{\textbf{Setting-All}} shows a reduction ranging from 5.2\% -- 29.1\% compared to \textit{\textbf{Setting-Common}}, a reduction ranging from 3.8\% -- 19.0\% compared to \textit{\textbf{Setting-NonImage}}, and a reduction ranging from 1.4\% -- 14.3\% compared to \textit{\textbf{Setting-Image}}. Among these, the reduction range is the largest when compared to \textit{\textbf{Setting-Common}}, suggesting that common covariates serve as the baseline. 

This finding indicates that using both image and non-image information improves prediction accuracy over using only either information.
It also implies that information contained in non-image variables and image or color variables do not completely overlap, and they each provide some distinct contribution to the prediction of ``Likes''. In this setting, RF and XGB show the smallest errors compared to the linear models. Regarding RMSE reduction, XGB and RF show improvement of 47.4\% and 27.3\% compared to LM, respectively.

Between \textit{\textbf{Setting-NonImage}} and \textit{\textbf{Setting-Image}}, the performance results differ method by method. For instance, in terms of MAE, LM and LMM, which accounts for linear relationship, show better performance under \textit{\textbf{Setting-NonImage}} than \textit{\textbf{Setting-Image}}, whereas SVR, MLP, RF, and XGB show the vice versa. 
This suggests that image covariates may have complex underlying structures, in which case linear relationships can be overly restrictive and incapable of capturing such a structure, including interactions among covariates. In such cases, more flexible approaches, such as tree-based models, demonstrates their strengths.

With respect to the six methods, comparing LM and LMM for all settings first shows that there is a large gain in accuracy for LMM over LM. Given that LMM better considers the hierarchical structure, such an improvement provides suggestive evidence that a user-post hierarchical structure is indeed present in social media data. Second, across all four settings, RF and XGB yielded the smallest test error. This result shows that social media data for popularity prediction not only has hierarchical structure, but also further displays complex relationships between covariates that are better captured using tree-based models.

\begin{figure*}[h]
    \centering
    \includegraphics[width=0.55\textwidth]{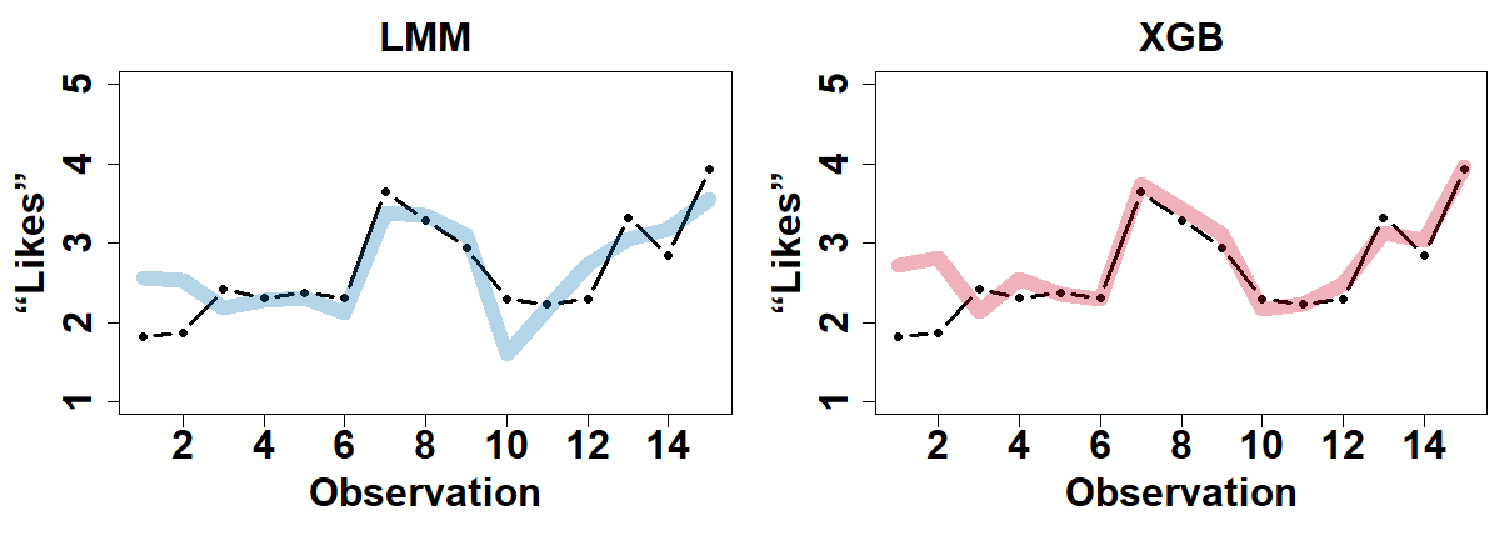}
    \caption{Actual and predicted ``Likes'' for a user. The connected black lines display actual ``Likes'', and the solid lines display predicted ``Likes'' from LMM (left) and XGB (right), respectively}
    \label{fig:user_pred}
\end{figure*}

We present consistent findings in Figure \ref{fig:user_pred}, which shows that XGB, a representative model that captures nonlinear correlation outperforms LMM, a representative model that captures linear correlation. In Panels (a) and (b), we present the actual values of ``Likes'' and the predicted ``Likes'' using LMM (left) and XGB (right) for a user. XGB displays an improvement in modeling the actual post popularity compared with LMM. 

Based on XGB results, which show the best performance, we further discuss the importance of each covariate in the ``Likes'' prediction. Since the prediction results of XGB are complex and difficult to interpret, we use `TreeSHAP' \citep{TreeSHAP} to obtain an interpretable measure of covariate importance. 
TreeSHAP is a method for estimating the SHAP (SHapley Additional ExPlanations, \cite{SHAP}) value of the tree-based model, where SHAP represents the relative importance of each covariate in the prediction. A larger SHAP value implies a higher importance of the covariate in the prediction. 

\begin{figure*}[h]
\centering
    \includegraphics[width=0.6\textwidth]{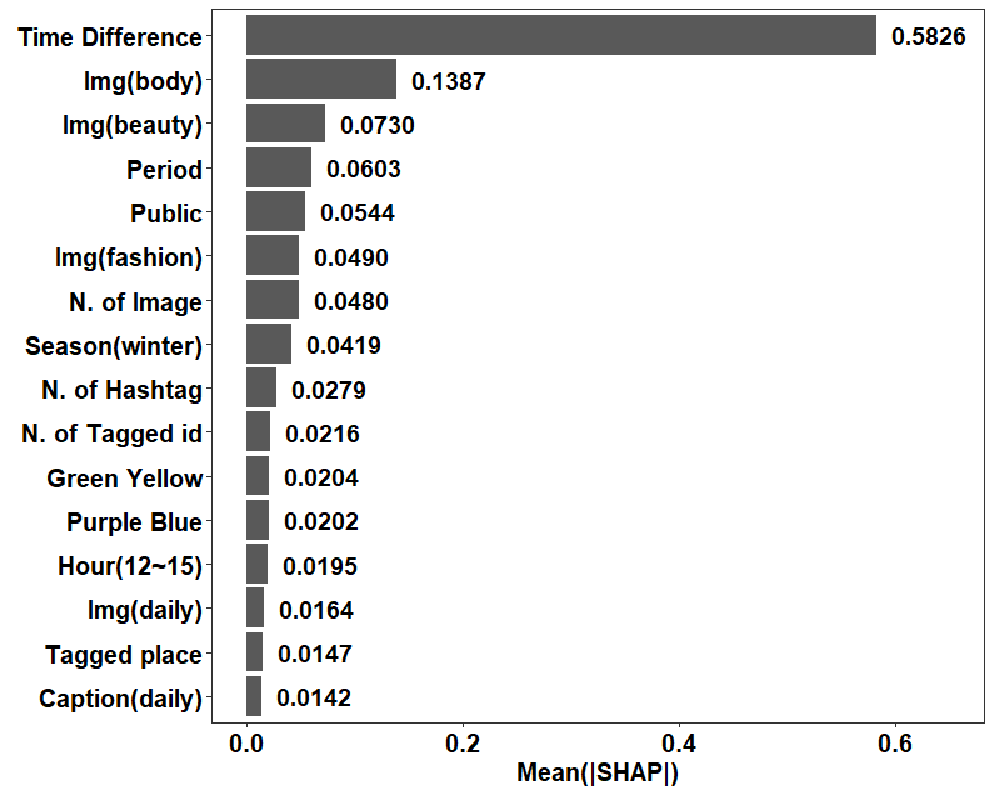}
    \caption{Covariate Importance Plot. Mean SHAP values are displayed for the top 16 covariates used in the XGB results except for user dummies}
    \label{fig:feature_importance} 
\end{figure*}

In Figure \ref{fig:feature_importance}, we display the top 16 covariates in the order of mean absolute SHAP values of the XGB results.\footnote{Each covariate has SHAP values as many as the number of observations.} 
The mean SHAP value for `Time Difference' is substantially larger compared to other covariates, followed by Image label topic `Body'. Image label topics are ranked high, suggesting that image information may play a non-trivial role in post popularity prediction. Within the Image label topics, `Body', `Beauty', and `Fashion' were of relatively higher importance, whereas `Daily' and `Food' rank much lower.\footnote{`Food' is not ranked among the top 16 covariates. For a full set of mean SHAP values for all covariates and a more detailed discussion on the SHAP values for each covariate, see Appendix \ref{Appendix:SHAP}.} This may reflect the nature of the social media platform, where visually more salient topics affect the popularity of a post more than less conspicuous topics. In terms of color, the representative color belonging to the `GY (green yellow)' and `PB (purple blue)' were more important for ``Likes'' than other color categories. Overall, Image label topics seem to be of higher importance than caption topics, which is consistent with social media platforms emphasizing visual content.

\section{Conclusion and future work}\label{sec6:conclusion}

In this study, we explore the intricate characteristics of social media data in various aspects.
First, we compare the performance of six models that could address complex data structures with a linear model as a baseline in predicting the popularity of image-based social media posts. Among the considered models, namely, Linear Mixed Model, Multi-layer Perceptron, Support Vector Regression, Random Forest, and XGBoost, we find that XGBoost outperforms the rest. 
Next, we investigate an easy-to-use variable construction method, utilizing Google Cloud Vision API, which limits the discretion of the researcher. We further process the resulting labels to generate a class of variables that summarizes the image-level information of a given post and show that it improves the model fit. These variables also provide interpretable results. Overall, our proposed framework is practical and capable of capturing useful information for social media data analysis. Our findings confirm that accounting for the complex structure inherent in the data is important for predicting post popularity.

Our work suggests several meaningful avenues for future research. First, our research is only utilizing image and text information, but it would be beneficial to investigate further how to consider video information as well. As video formats are also gaining wider interest, developing models of video post popularity are in high demand. Second, the prediction of alternative popularity measures may provide additional insights into related fields. Liking may not always be the most relevant behavior to popularity, since underlying follower satisfaction or attitudes are not fully accounted for in this metric. Therefore, additionally employing alternative measures of popularity, such as follower sentiments in the comments would provide a more complete picture regarding post/user popularity. 
Finally, besides popularity prediction, another dimension that may be useful for practitioners and researchers is uncovering the causal factors of social media post popularity. While social media platforms expose posts to users by endogenously and strategically defined targeting functions or algorithms, disentangling the causal factors is an extremely complex but important task. We leave these questions to future work.

\appendix
\section*{Appendix}

\setcounter{table}{0}
\renewcommand{\thetable}{\thesection.\arabic{table}}

\setcounter{figure}{0}    
\renewcommand\thefigure{\thesection.\arabic{figure}}    

\section{Sampling process}\label{app:usersampling}

In this paper, we sought to leverage text and image data of social media data for the purpose of predicting the number of ``Likes''. However, to ensure the quality and relevance of the data used in our study, we implemented a careful selection process.

Our sample consists of influencers with a public account, who among their latest 100 posts, have uploaded a minimum of five sponsored or advertisement posts. This is determined by hashtags, such as \#AD or \#Sponsored. We further restrict our sample to those who have uploaded at least 100 posts at the time of crawling. Instead of relying on random users' public data, which often includes individuals who may not frequently post images or share poor quality and meaningless content, we adopted a targeted approach by focusing on active influencers. By selecting users who demonstrate a consistent pattern of posting visually appealing and engaging photos, we aimed to capture a representative sample of users whose content and audience interactions are more likely to yield meaningful insights for predicting the number of ``Likes''. This approach allowed us to enhance the reliability and validity of our analysis, ensuring that the data used accurately reflects the dynamics and factors influencing user engagement on social media. This sample selection process allowed us to focus on users who provide regular uploads, post meaningful content, and tend to post diverse topics to actively engage the audience on the platform.

Additionally, we wanted that the number of observations for each user is large for several reasons. First, in the popularity prediction task, it is critical to consider individual user effect because many user-specific unobservable factors, such as user engagement habits, content style, creativity and degree of popularity of the user, may substantially influence post popularity. Ensuring enough observations for each user enables us to disentangle the user-specific fixed effect. Next, examining a substantial number of posts per user allows us to capture a more comprehensive picture of users' posting behavior and content patterns. Additionally, analyzing a larger sample size helps mitigate the influence of outliers or temporary fluctuations, providing a more accurate representation of their typical posting style and engagement levels.

Specifically, we conducted the following procedure for sample construction: 
\begin{enumerate}
    \item To select an initial user, we search for the hashtag `$\#$Sponsor' and choose the user who appears at the top of the search feed. Note that only the public account users will appear in the search feed. This user is assigned as ``parent user''.
    \item We extract the users who are present in the ``parent user'''s following list. These users are assigned as ``children users''.
    \item Among the ``children users'', we select users who meet the criteria mentioned above: (1) having more than 100 posts and (2) having more than five sponsored/advertisement posts among their most recent 100 posts.
    \item These selected users are assigned as ``parent user''.
    \item We repeat 2, 3, and 4.
\end{enumerate}

We collected data of 40 users, and to give the same information weight to the selected users, we crawled the 100 most recent posts for each user. Although we collected 100 post information for all users, the total number of uploaded images may vary for each user, as each post can contain anywhere from one to ten images. The data crawling period is between 6 February 2022 and 24 March 2022, with the resulting analysis sample of 40 users, 3,807 posts, and 13,774 images. From 4,000 collected posts, those consisting only video contents without any image are excluded.

There are a few aspects of the sampling process that need to be mentioned. First, in our data collection process, usernames are not acquired. Instead, individuals are assigned randomized identification numbers for the purpose of anonymity and confidentiality. User individuals effects are addressed by using dummy variables for each user. Second, restricting the user selection pool to the following account list of the previously chosen user offers a realistic and practical approach for data collection. This is because users tend to follow accounts with similar interests or content, making it more likely to identify individuals who actively engage with sponsored posts.

\section{Seeded-LDA}\label{Appde}

\subsection{The settings of Seeded-LDA}\label{app:slda_param}

The LDA's hyperparameter is the parameter of the prior distribution of words and documents. In general, $\alpha$ is a parameter of the document distribution, and $\beta$ is a parameter of the word distribution. The tuning grid was set to 0.05, 0.1, 0.5, 1, 5, and 10, and the hyperparameter was selected based on the logarithmic ratio among the measurements of coherence. For caption topics, ($\alpha$,$\beta$) was set to (10,10), and for Image label topics, ($\alpha$,$\beta$), the value of (10,0.1) was selected. The seed words for each topic are listed in Table \ref{tab:seedword}.

\begin{table*}[h]
\caption{The list of seed words used for each topic in performing Seeded-LDA for caption and image labels is based on frequency }
\label{tab:seedword}
\centering
\small
\subtable[Seed words for caption topics. Proper noun is starred]{ 
\footnotesize
    \begin{tabular}{|c|c|c|c|c|}
    \hline
    \textbf{Event} & \textbf{Beauty} & \textbf{Health} & \textbf{Fashion} & \textbf{Daily} \\
    \hline
    Event & Olive Young$^*$ & Diet & Hoodie & Date \\
    First-come & Lipstick & Collagen & Styling & Weather \\
    Point & Eyelashes & Lactobacillus & Design & Daily Life \\
    Promotion & Makeup & Stress & Style & Highlight \\
    Experience Team & Collagen & Probiotics & Daily Look & Update \\
    Discount Rate & Mascara & Healthy & Outerwear & Year-end \\
    Repurchase & Mask Pack & Dietary Fiber & Pants & Jeju \\
    Service & Lifting & Vitamin & Shoes & Shooting \\
    Discount & Inner Beauty & Balance & Jacket & Video \\
    Comment & Mist & Food & Padded Jacket & Birthday \\
    Certification & Pouch & Exercise & Cotton & YouTube \\
    Announcement & Clinic & Delicious & Collection & Holidays \\
    Draw & Essence & Health & Length & Foodie \\
    Contact & Acne & Take care of & Knit & Vaccine \\
    Winner & Skincare & Health Functional Food & Wear & Youtube Live \\
    Recipient & Cleanser & Diet & Put on & Cafe \\
    Gift & Anti-aging & Weight Loss & Bag & End-of-year \\
    Participation & Trouble & Supplement & Hair & Healing \\
    Bulk Purchase & Apply & Digestion & Shopping & Movie \\
    Winning & Procedure & Body Fat & Attire & New Year \\
    \hline
    \end{tabular}
}

\hspace{5mm}
\subtable[Seed words for image label topics]{
\footnotesize
    \begin{tabular}{|c|c|c|c|c|}
    \hline
    \textbf{Fashion} & \textbf{Food} & \textbf{Body} & \textbf{Beauty} & \textbf{Daily} \\
    \hline
    Sleeve & Food & Human Body & Lipstick & Flash Photography \\
    Fashion & Noodle & Neck & Cosmetics & Travel \\
    Trousers & Baked Goods & Shoulder & Nail Care & Leisure \\
    Jacket & Cooking & Thigh & Body Jewelry & Christmas Decoration \\
    Overcoat & Pasta & Muscle & Eye Shadow & Shopping \\
    Style & French Fries & Chest & Eye Liner & Entertainment \\
    \hline
    \end{tabular}
}
\end{table*}

\subsection{Sensitivity of the number of topics}\label{app:slda_topic}

Seeded-LDA can be sensitive to the selection of several pre-determined topics and seed words. To investigate the impact of these factors, we conduct experiments to evaluate topic diversity of our predefined 5 topics, varying the number of topics from 2 to 5. For each value of $k$, we conduct $5 \choose k$ experiments.

In each experiment, $k$ seed sets are selected from a pool of predefined 5 seed sets used in Table~\ref{tab:seedword}. We evaluate topic diversity in each experiment using Jaccard similarity. For two word sets $T_i$ and $T_j$, Jaccard similarity is defined as:
\begin{align*}
JS(T_i,T_j) = \frac{|T_i \cap T_j|}{|T_i \cup T_j|}
\end{align*}
For $1 \leq i < j \leq k$, where $k$ is the number of topics, topic diversity is calculated by averaging pairwise Jaccard similarities between topics:
\begin{align*}
\text{Topic Diversity} = \frac{2}{k(k-1)} \sum_{i<j} JS(T_i,T_j)
\end{align*}
Topic diversity measures the extent to which the topics captured by captions and image labels represent diverse aspects of the data and quantifies the distinctiveness between topics. 

\begin{figure}[h]
\subfigure{
\centering
\includegraphics[width=.48\textwidth]{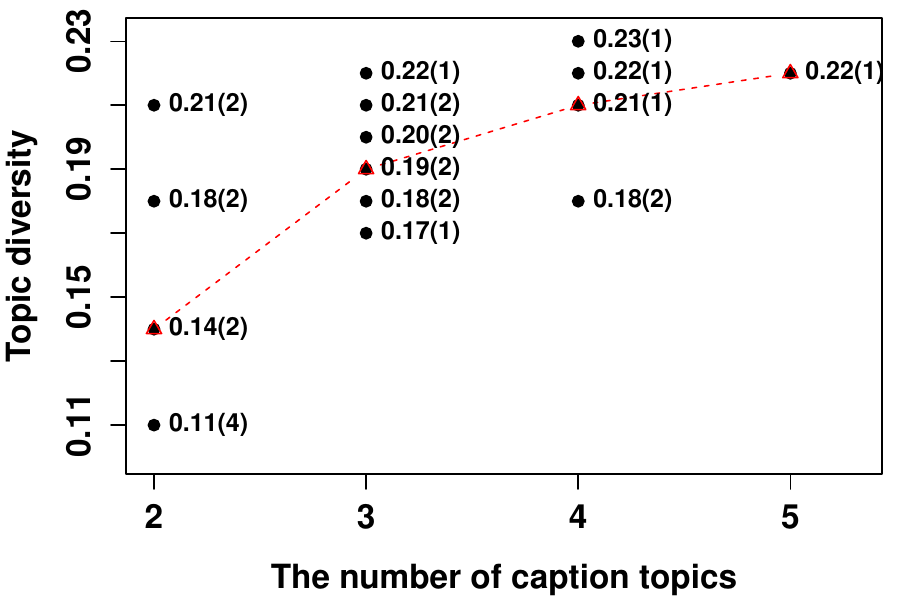}
}
\subfigure{
\centering
\includegraphics[width=.48\textwidth]{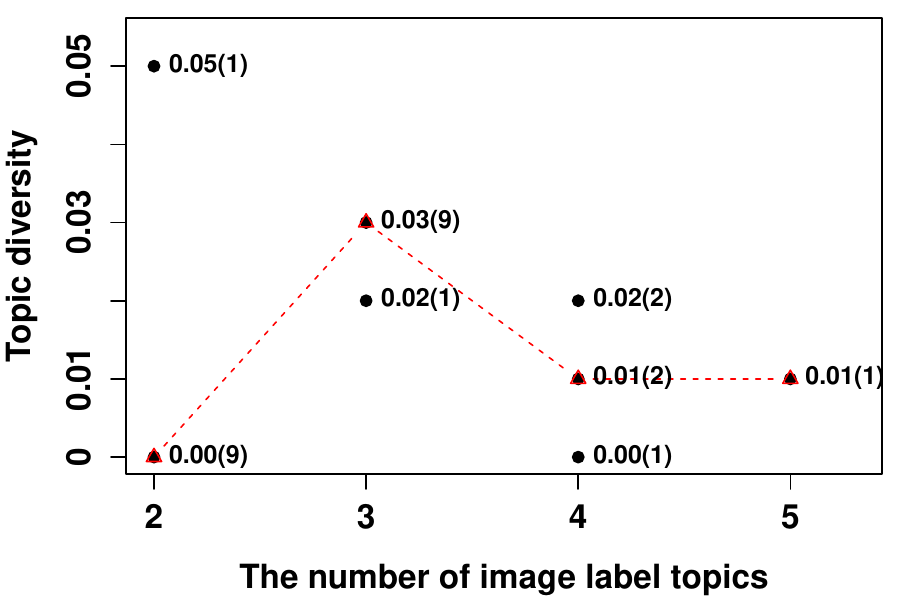}
}
\caption{Topic diversity of our seed sets. The numbers in parentheses denote the count of overlapping points in the value of topic diversity. The red line connects the median values for all possible combinations of specific topic numbers}
\label{fig:top_diversity} 
\end{figure}

The value of topic diversity for $k$'s 2 to 5 are illustrated in Figure~\ref{fig:top_diversity}. For captions, employing larger topic numbers results in higher overlap between the topics (left panel). This could be attributed to the common terms regularly used by users, causing them to frequently appear in various topics. As an example, the term ``cosmetic'' often appear in both event and sponsorship posts. Image labels, by contrast, consistently exhibit a lower overlap, indicating that they encompass a broader range of information embedded within the data (right panel).

\subsection{Sensitivity of seed selection}\label{app:slda_seed}

To explore the sensitivity of our results to the chosen seed word sets, we construct an alternative seed word sets using LDA for comparison, which we refer to as `contrastive set'. We first performed LDA with five topics. Then, for each topic, we identify the words with the highest information gain. 
The information gain of a word $w$ to a topic $T$ is calculated as:
\begin{align*}
IG(T,w) = H(T) - H(T|w),
\end{align*}
where $H(T)$ represents the entropy of the topic, and $H(T|w)$ represents the conditional entropy of the topic given the presence of word $w$. 
We then select the words with the highest information gain for each topic as the contrastive seed words, removing the words that overlapped across topics. To maintain consistency with the original seed set size, we chose 20 and 6 final seed words per topic for captions and image labels, respectively.
The selected constrastive seed word sets are exhibited in Table~\ref{tab:seedword_ig}.

\begin{table*}[h]
\caption{The list of seed words selected by Information gain}
\label{tab:seedword_ig}
\centering
\small
\subtable[Seed words for caption topics. Proper noun is starred]{ 
\footnotesize
    \begin{tabular}{|c|c|c|c|c|}
    \hline
    \textbf{Event} & \textbf{Beauty} & \textbf{Health} & \textbf{Fashion} & \textbf{Daily} \\
    \hline
    Event & Hair & Calories & Coat & Cafe \\
    Brand & Lifting & Stamina & Length & Friends \\
    Order & Acne & Nutrition Supplement & Shoes & YouTube \\
    Bulk Purchase & Pores & Extract & Joggers & Day \\
    Membership & Endurance & Organic & Fabric & Fine Dust \\
    DM (Direct Message) & Soothing & Weight Gain & Denim & Coffee \\
    Website & Beauty & Health & Mustard & Vaccine \\
    Video & Olive Young & Full stomach & Jeans & Christmas \\
    Service & Makeup & Empty stomach & Mink & Healing \\
    New Product & Exfoliate & Diet & Pants & Beer \\
    Winner & Trouble & Food Additives & Outfit & Parents \\
    Online & Inner Beauty & Fructooligosaccharide & Wearing & Everland$^*$ \\
    Renewal & Toner & Eco-friendly & Dress & Brunch \\
    Repurchase & Apply & Fasting & Shirt & Vacation \\
    Online Mall & Moisture & Weight & Walker & Movie \\
    Coupon & Warm Tone & Probiotics & Knit & Year-end \\
    Comment & Clinic & Exercise & Hoodie & Glamping \\
    Store & Pouch & Food & Sneakers & Daily Life \\
    Opportunity & Mask Pack & Healthy & Shopping & Update \\
    Experience Team & Skincare & Recovery & Outerwear & Birthday \\
    \hline
    \end{tabular}
}
\hspace{5mm}
\subtable[Seed words for image label topics]{
\footnotesize
    \begin{tabular}{|c|c|c|c|c|}
    \hline
    \textbf{Fashion} & \textbf{Food} & \textbf{Body} & \textbf{Beauty} & \textbf{Daily} \\
    \hline
    Sleeve & Food & Waist & Jewelry & Sky \\
    Denim & Cuisine & Knee & Lipstick & Landscape \\
    Sportswear & Dish & Thigh & Makeover & Tourism \\
    Fashion Design & Tableware & Human Leg & Beauty & Travel \\
    Blazer & Comfort Food & Neck & Cosmetics & Leisure \\
    Uniform & Recipe & Chin & Lip Gloss & Vacation \\
    \hline
    \end{tabular}
}
\end{table*}

\begin{figure}[h]
\subfigure{
\centering
\includegraphics[width=.48\textwidth]{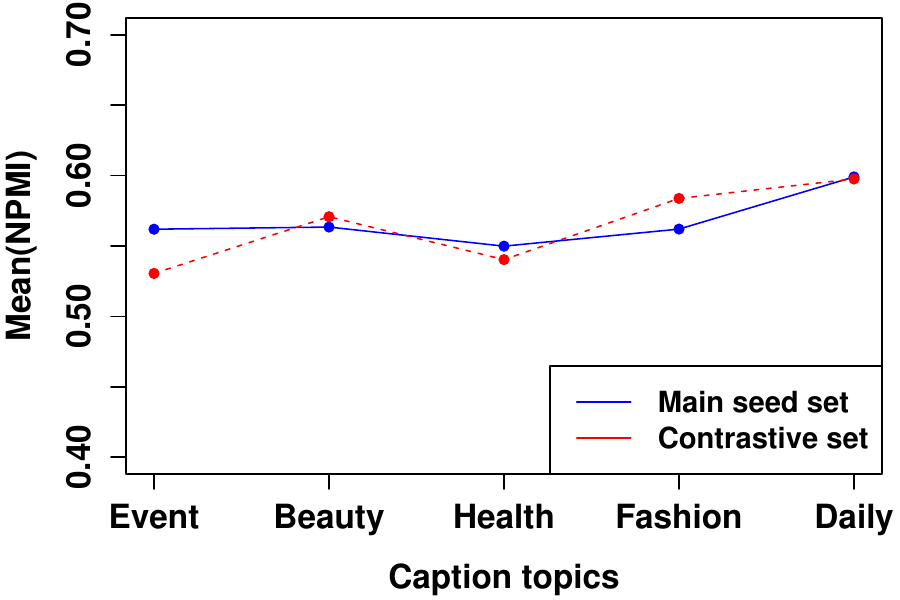}
}
\subfigure{
\centering
\includegraphics[width=.48\textwidth]{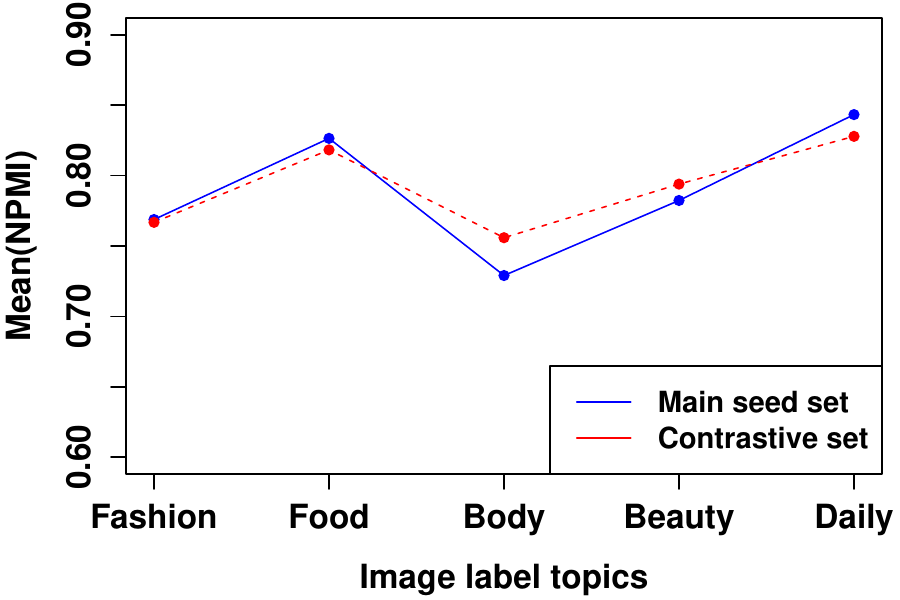}
}
\caption{Topic coherence of two seed sets}
\label{fig:top_coherence} 
\end{figure}

To evaluate the effectiveness of seed word sets, we assess the coherence of topics generated using these sets. 
We employ the average Normalized Pointwise Mutual Information (NPMI) as the evaluation measure.
The average NPMI of a topic is evaluated across all representative word pairs within the given topic:
\begin{align*}
\text{NPMI}(T) = \frac{2}{N(N-1)} \sum_{i=1}^{N-1} \sum_{j=i+1}^{N} \frac{\log_2 \frac{P(r_i(T),r_j(T))}{P(r_i(T),r_j(T) )P(r_j (T))}}{-\log_2 P(r_i (T),r_j (T))}.
\end{align*}
where $N$, $r_i (T)$ for $i=1, \cdots, N$, and $P$ denote the number of representative words, the $i-$th representative word of topic $T$, and the frequency of the input arguments appearing across the documents. 
The representative words $r_i(T)$ of a topic $T$ are selected based on their high relevance as proposed by \cite{ldavis}. The relevance $r(\cdot | \ \cdot)$ is defined as follows:
\begin{align*}
r(w|T) = \lambda \cdot P(w|T) + (1-\lambda) \cdot \frac{P(w|T)}{P(w)}.
\end{align*}
Figure~\ref{fig:top_coherence} displays the average NPMI score of each topic achieved by the proposed seed word sets and the contrastive seed word sets.
This analysis reveals a similar trend for both our proposed seed word sets and the contrastive sets. This indicates that the topic coherence, as measured by NPMI score, exhibits relatively low sensitivity to the specific choice of seed words in this dataset.

\section{Method settings}\label{Appendix:tune_result} 

\subsection{Linear Mixed Model}

In our analysis, the covariates that yield singular results when modeled in the simple Linear Mixed Model (Linear Mixed Model with only one covariate), are set to have fixed effects only. These covariates include `Period', `Weekdays', `Hour', `Season', `Topic (event)', `G (green)', and `BG (blue green)'. In addition to these seven covariates, the `Time difference' is also modeled to have only a fixed effect owing to computational issues. The remaining variables are modeled to have both random and fixed effects.

\subsection{Support Vector Regression}

In our analysis, kernel Support Vector Regression is used to account for structures of covariates in a more flexible way. Specifically, the original $p$-dimensional covariates are mapped to infinite dimensional space by the function $\phi: \mathbb{R}^{p} \rightarrow \mathbb{R}^{\infty}$ where
\begin{equation}
\label{eq:radial}
   \langle  \phi(\mathbf{x}_1), \phi(\mathbf{x}_2) \rangle  = \exp \left( -\gamma \vert\vert \mathbf{x}_1 - \mathbf{x}_2 \vert\vert _2^2\right),
\end{equation}
and $\phi(\mathbf{x})$s are used as infinite- dimensional covariates replacing the $p$-dimensional covariates $\mathbf{x}$. The prediction of $y$ given the covariate vector $\mathbf{x}$ becomes as follows:

\begin{align}
\label{obj:svr}
    \hat{y} &= \langle \hat{\beta}_1, \phi(\mathbf{x}) \rangle + \hat{\beta}_0, \nonumber \\
    (\hat{\beta}_0, \hat{\beta}_{1}) &= \underset{(\beta_0, \beta_{1})  \in \mathbb{R} \times \mathbb{R}^{\infty} } {\arg \min} \frac{1}{2} ||\beta_{1}||_2^2 + C \sum_{i}  (\xi_{i} + \xi_{i}^{*})  \nonumber \\
\text{subject to} \; &\left(\langle \beta_1,  \phi(\mathbf{x}_{i}) \rangle + \beta_0\right) - y_{i} \leq \epsilon + \xi_{i}, \nonumber\\
& y_{i}-\left(\langle \beta_1, \phi( \mathbf{x}_{i} ) \rangle + \beta_0\right)  \leq \epsilon + \xi_{i}^{*}, \nonumber \\ 
& \text{and} \quad \xi_{i}, \ \xi_{i}^{*} \geq 0.
\end{align}

\subsection{Parameters settings}

SVR involves three tuning parameters: $\epsilon$, $C$ in \eqref{obj:svr}, and $\gamma$ in \eqref{eq:radial}. $\epsilon$ is the allowed range of residuals around the regression line, $C$ is the regularization parameter of the residuals that exceed the allowed range $\epsilon$, and $\gamma_{\text{svr}}$ controls the smoothness of the radial kernel.

RF and XGB share several common tuning parameters related to their tree structures, namely, the total number of trees ($B$), maximum depth of a tree ($d$), fraction of covariates to be used for each split of a tree ($v$), fraction of subsamples to be used in building each tree ($s$), and minimum number of observations that each terminal node should contain ($m$). 

XGB has additional tuning parameters $\rho$ and $\gamma_{\text{XGB}}$ that control overfitting: $\rho$ is the learning rate of gradient boosting, and $\gamma_{\text{XGB}}$ is the minimum loss reduction required for a tree split. Here, $\gamma_{\text{XGB}}$ plays the same role as $\gamma$ described in \ref{sub:xgb} by penalizing a large tree. Another regularization parameter, $\lambda$ in \ref{sub:xgb} is set to $1$ by default.

MLP involves two parameters for the hidden layer: hidden layer size $h$ and activation function. We employ Rectified Linear Unit (ReLU) as the activation function for $h=5$. For optimization, MLP selects the optimizer and the learning rate $\rho$. We opt for the Adam optimizer, initializing the learning rate at $\rho=0.001$ and adaptively adjusting it during training. We use a batch size of full data to ensure a fair comparison with other methods in terms of cross validation and train the model for 500 epochs.

The R packages `e1071', `ranger', and `XGBoost' are used for SVR, RF, and XGB, respectively. The python package `keras' are used for MLP. The tuning parameter settings are presented in Table \ref{tab:tune_result}.

\begin{table}[!h]
\caption{ Tuning parameter settings}
\label{tab:tune_result}
\small
\centering
\begin{tabular}{ccccccc}
\toprule
Method & Parameter & Description & {All} & {Non-image} & {Image} & {Common} \\
\cmidrule(l{0pt}){1-3} \cmidrule(l){4-7}
\multirow{3}{*}{SVR} & $\epsilon$ & Range of residuals & 0.2 & 0.2 & 0.2 & 0.3 \\
& $C$ & Regularization of range violation  & 85 & 95 & 95 & 100 \\
& $\gamma_{\text{SVR}}$ & Radial kernel smoothness & 0.0008 & 0.0009 & 0.003 & 0.1 \\
\cmidrule(l{0pt}){1-7} 
\multirow{5}{*}{MLP} & $h_1$ & First hidden layer size & 1024& 1024 &1024& 1024\\
& $h_2$ & Second hidden layer size & 128& 64 &16& 128\\
& $h_3$ & Third hidden layer size & 128 & 512 &16& 32\\
& $h_4$ & Fourth hidden layer size & 32 & 32 &128 & 256\\
& $h_5$ & Fifth hidden layer size & 1024 & 1024 &1024& 1024\\
\cmidrule(l{0pt}){1-7}
\multirow{5}{*}{RF} & $B$ & Total number of trees & 500 & 3000 & 1000 & 1000 \\
& $d$ & Maximum tree depth & 40 & 40 & 40 & 40 \\
& $v$ & Covariates fraction & 1 & 0.6 & 0.6 & 0.7\\
& $m$ & Minimum node size & 10 & 10 & 10 & 15\\
& $s$ & Subsample fraction & 1 & 1 & 1 & 0.6\\
\cmidrule(l{0pt}){1-7} 
\multirow{7}{*}{XGB} & $B$ & Total number of trees & 2000 & 1000 & 1000 & 1000\\
& $d$ & Maximum tree depth & 10 & 10 & 10 & 10 \\
& $v$ & Covariates fraction & 0.8 & 1 & 0.9 & 0.8 \\
& $m$ & Minimum node size & 5 & 5 & 5 & 10\\
& $s$ & Subsample fraction & 0.6 & 0.6 & 0.6 & 0.6\\
& $\rho$ & Learning rate & 0.01 & 0.01 & 0.01 & 0.01\\
& $\gamma_{\text{XGB}}$ & Minimum loss reduction & 0.005 & 0.005 & 0.005 & 0.005\\
\bottomrule
\end{tabular}
\end{table}

\section{TreeSHAP results}\label{Appendix:SHAP}

\begin{figure*}[htbp]
\centering
    \includegraphics[width=0.8\textwidth]{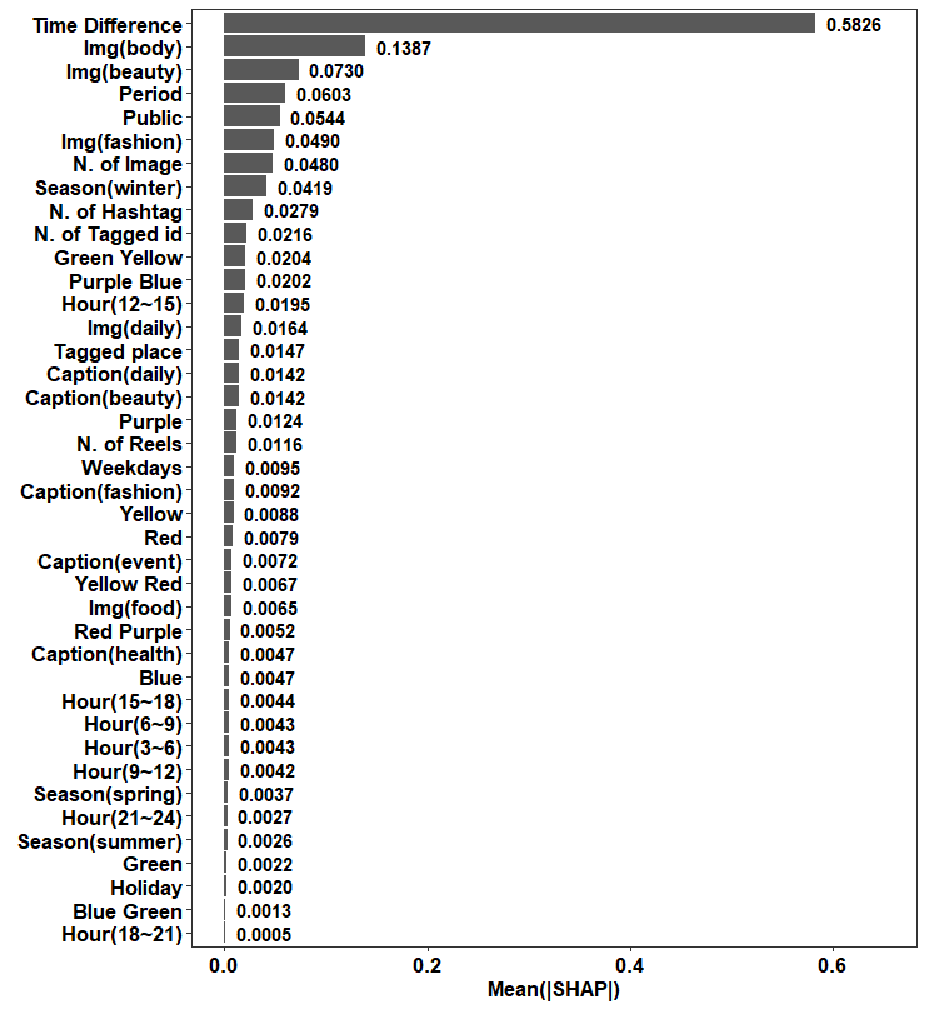}
    \caption{Covariate Importance Plot. Mean SHAP values are displayed for each covariate used in the XGB results, except for user dummies}
    \label{fig:treeshap_plot_1}
\end{figure*}

\begin{figure*}[htbp]
\centering
    \includegraphics[width=0.8\textwidth]{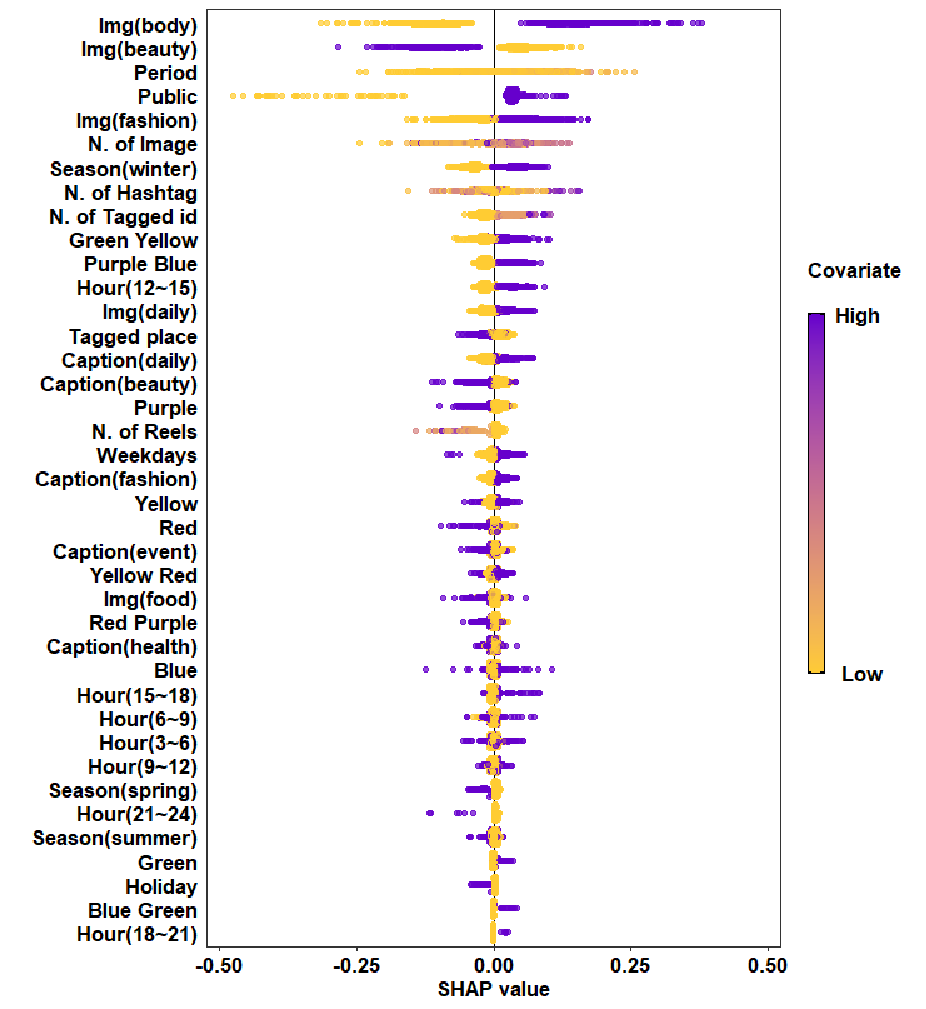}
    \caption{Covariate Summary Plot. For each covariate, the corresponding SHAP values are visualized, except for user dummies and `Time Difference' for visibility}
    \label{fig:treeshap_plot_2} 
\end{figure*}

\begin{figure*}[htbp]
\centering
    \includegraphics[width=1\textwidth]{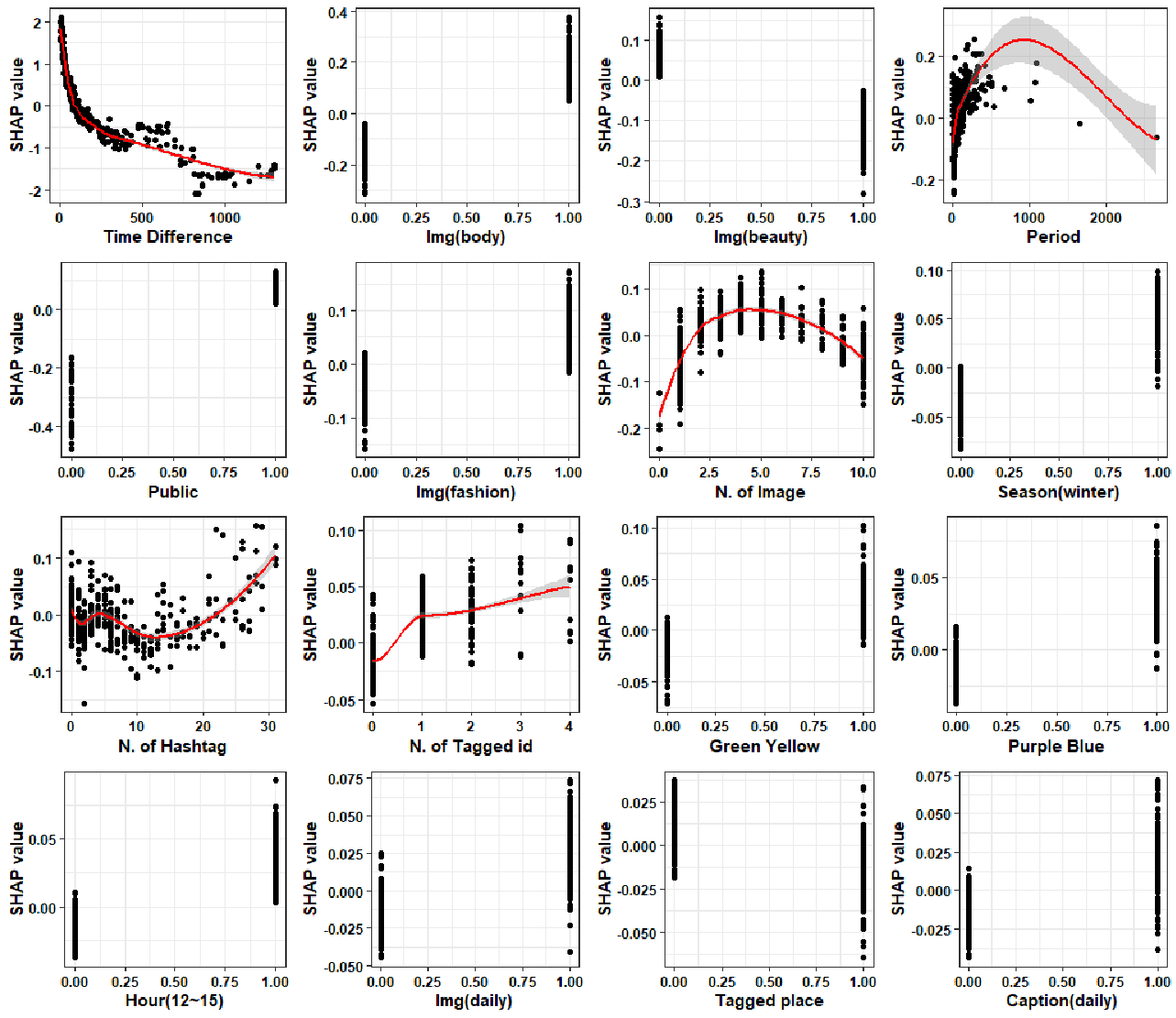}
    \caption{Dependence Plot for the top 16 covariates}
    \label{fig:treeshap_plot_3}
\end{figure*} 

Figure \ref{fig:treeshap_plot_1} displays the mean absolute SHAP values of all covariates in the XGB result. Figure \ref{fig:treeshap_plot_2} shows the observation-wise SHAP values, in which the covariate value of each observation is represented by a color gradient. The closer the color is to dark purple, the greater the value of each covariate.

Figure \ref{fig:treeshap_plot_3} shows the SHAP dependence plot for the top 16 covariates in order of SHAP values. This plot shows a scatterplot of the SHAP values against the covariate values, along with the trend line for all observations.

The top left corner presents the plot for `Time Difference'. Since we considered the latest 100 posts for a given user in the data collection process, most observations are posts uploaded within a year. The trend line of the plot indicates that the older the post, the greater the decline in the importance of the covariate for prediction of ``Likes'' until around the first 100 days. This suggests that people may not react to old posts because they are mainly interested in new posts on social media. Therefore, considering such characteristics of social media, where new posts are constantly uploaded, it can be said that it is important to use `Time Difference' information.

Another time lapse variable, `Period', was also important, displayed in the rightmost corner. Most users in our dataset upload new posts within 250 hours (approximately 10 days) of uploading the previous post. For `Period' plot, when the upload cycle of posts is short, the influence on ``Likes'' is low, but as the posting cycle gradually increases, the importance tends to increase. That is, when uploading a post, it may be desirable for the user to upload the post with a certain time interval rather than a cycle that is too short for higher popularity. 

Another variable, `N. of Image' refers to the number of images uploaded in a single post. There are various distributions of `N. of Image', starting from 0 (when only reels are posted). An inverse U-shaped relationship is observed, which suggests that both too few images and too many images in a single post are not desirable, and that there may exist an optimal number of images per post. 

Among Image label topics, the covariate importance was positively higher when there were `Body' and `Fashion' topics in the post than when there were no such topics, but it was negatively higher when `Beauty' topic was included in the post.

\section{Comparison of covariate construction approaches}\label{app:compare}

In this section, we present a comparative analysis on covariate construction methods using `User' and `Time difference' as baseline covariates. Based on the previous section's findings, XGBoost is chosen as the prediction model due to its effectiveness in describing image-based social media data.
We evaluate the performance of our proposed covariate construction approaches alongside other methods in image covariate (image and representative color), and non-image covariate (caption) construction, respectively.
More precisely, we focus on contrasting with neural network-based methods, which is known to display strong performance but often challenging to interpret. Previous studies have utilized sequential steps of multimodal fusion processes, employing neural networks (NNs), to obtain unique overall vector representation for captions or images (e.g., \cite{WANG2023101490}). The resulting vector representation encapsulates fused meanings from various multimodal sources, including text from post captions, image captions, and additional features. However, this compromises the interpretability of the resulting vector representation.
The comparison results are exhibited in Table~\ref{tab:comparison}.

\subsection{Image covariate}

\begin{table}[h]
    \centering
    \caption{Test errors for different variable construction methods. 
    For each covariate, the starred value represents the minimum error. The results from our proposed approaches are denoted in italic.}
    \label{tab:comparison}
    \belowrulesep = 0pt
    \aboverulesep = 0pt
    \begin{tabular}{ccccc}
    \toprule
    & & & \multicolumn{2}{c}{XGBoost} \\
    \cmidrule(l{0pt}){4-5}
    & & & RMSE & MAE  \\
    \cmidrule(l{0pt}){1-5}
    \multirow{5}{*}{Image} & \multicolumn{1}{|c|}{\multirow{3}{*}{Content}}  &  Image class  & 0.5770 & 0.4357 \\
    & \multicolumn{1}{|l|}{}& Image caption topic & 0.6447 & 0.4756\\
    & \multicolumn{1}{|l|}{}& \textit{Image label topic} & \hspace{0.5mm} \textit{0.5557}$^{*}$ & \hspace{0.5mm} \textit{0.4142}$^{*}$ \\
    \cmidrule(l{0pt}){2-5}
    & \multicolumn{1}{|c|}{\multirow{2}{*}{Color}} & HSV space & 0.6360 & 0.4758\\
    & \multicolumn{1}{|l|}{}& \textit{Munsell space} & \hspace{0.5mm} \textit{0.6283}$^{*}$ & \hspace{0.5mm} \textit{0.4741}$^{*}$\\
    \cmidrule(l{0pt}){1-5}
    \multicolumn{2}{c|}{\multirow{3}{*}{Caption}} & Caption deep feature & 0.6428 & 0.4813\\
    & \multicolumn{1}{l|}{}  & Sentiment score & 0.6254  & 0.4638 \\
    & \multicolumn{1}{l|}{} & \textit{Caption topic} & \hspace{0.5mm} \textit{0.5806}$^{*}$ &\hspace{0.5mm}  \textit{0.4382}$^{*}$\\
    \bottomrule
    \end{tabular}
\end{table}

For image covariate, we employ two comparison methods to the proposed Google Vision API-based approach, each based on deep convolutional neural networks (DCNNs) and Bootstrapping Language-Image Pre-training (BLIP), referred to as `Image class', and `Image caption topic', respectively, with our proposed approach denoted by `Image label topic'.
Convolutional neural network-based models demonstrate outstanding performance in image processing. Thus, we employ comparison methods leveraging DCNNs and utilized their output image categories, following approaches outlined in works such as \cite{huang2017towards}, \cite{ hessel2017cats}, \cite{chen2016micro}, and \cite{overgoor2017spatio}. Specifically, we employ ResNet-50 architecture \citep{resnet} on an image and extract the resulting category with the highest probability. This process generates a binary vector of 700 dimensions, in which each element is an indicator of its associated image category. In the resulting vector, the element corresponding to an image category is encoded as 1, if the given image belongs to that category.
We then sum the binary vectors of images per post to form the post-level image covariate. Given that our approach involves extracting word labels from a given image, we also include performance comparison with a method that initially generates a caption from an image, followed by applying LDA on the caption, as proposed in \cite{8622461}. For generating image caption, we utilized BLIP \citep{BLIP} which is a convolutional neural network-based (CNN) approach.

For color covariate, we present a comparison of the prediction performance between Hue-Saturation-Value (HSV) color features, as used in \cite{7903630}, and the Munsell space employed in this study, referred to as `HSV space' and `Munsell space', respectively. To construct representative colors based on HSV color space, which is a color space commonly used in image analysis tasks, we first convert the RGB values of the image to the HSV. Subsequently, the converted HSV values are assigned to one of eight discrete classes: black/white, blue, cyan, green, yellow, orange, red, and magenta. Each color category is then summarized at the post-level, in the same manner as detailed in Section~\ref{subsubsec:img_color}.

Table \ref{tab:comparison} provides the test errors of various feature extraction methods. Among the compared image content processing methods, the proposed `Image label topic' exhibits the lowest error for both RMSE and MAE. Furthermore, `Image class' exhibits lower error compared to `Image caption topic'.
The results tend to improve as the image content is more directly utilized.
For instance, `Image label topic' utilizes more direct information than `Image class'. While `Image class' assigns a single category to an image, which can lead to an overly simplified representation, `Image label topic' methods utilize multiple labels extracted from a given image, resulting in a more direct and informative representation.
In addition, unlike `Image class', `Image caption topic' requires an additional step of converting the image into text. This indirect approach may be less suitable for social media platforms, where images are often used to convey a general impression rather than provide detailed information. In such contexts, the textual description of an image generated by `Image caption topic' might introduce irrelevant details that obscure the core message of the image, acting as noise in covariate construction.
These findings suggest that image posts on social media platforms carry information that can be represented solely by the image itself. Between two image color processing methods, `HSV space' and `Munsell space' show close results. The inherent similarity between HSV and Munsell spaces may contribute to the observed indifference in the outcome with respect to the chosen space.

\subsection{Non-image covariate}

For caption covariate, we compare the caption processing performance between the proposed Seeded-LDA approach, pre-trained neural network based model (BERT), and sentiment analysis algorithm (VADER), which are referred to as `Caption topic', `Caption deep feature', and `Sentiment score' in Table~\ref{tab:comparison}. In our comparison, BERT is utilized to explore an attention-based method.
Specifically, the caption covariates are extracted leveraging BERT, as employed in \cite{ding2019social}. We additionally compare caption features extracted through sentiment analysis using the VADER algorithm \citep{VADER}, as employed prior research in popularity prediction (e.g., \cite{keneshloo2016,SAEED2022116496}). The VADER outputs sentiment scores for 4 emotional categories--positive, negative, neutral, and compound--with the scores ranging between 0 and 1 where higher values represents stronger emotional intensity.

Among the caption processing methods, `Caption topic' achieves best results than both `Caption deep feature' and `Sentiment score'.
While both caption topic and caption deep feature represent textual content, caption deep feature exhibits a considerably higher dimensionality (768) compared to caption topic (5). This high dimensionality may lead to performance degradation.
Furthermore, `Sentiment score' showed limited results, which might stem from the inherent insufficiency of sentiment analysis.
Captions often convey richer information than sentiment alone.

\bmhead{Acknowledgments}
Dahyun Jeong and Yunjin Choi was supported by the National Research Foundation of Korea (NRF) grant funded by the korea government (MSIT, No. NRF-2022M3J6A1084845).

\bmhead{Conflict of interest}
On behalf of all authors, the corresponding author states that there is no conflict of interest.

\bibliography{reference}






\end{document}